\PassOptionsToPackage{table}{xcolor}
\documentclass[10pt,twocolumn,letterpaper]{article}
\usepackage[pagenumbers]{cvpr}
\makeatletter
\renewcommand{\LenToUnit}[1]{\strip@pt\dimexpr#1\relax}
\makeatother

\usepackage{nicefrac}
\usepackage{microtype}
\usepackage{multirow}
\usepackage{siunitx}
\usepackage{makecell}
\usepackage{threeparttable}
\usepackage{tabularx}
\usepackage{placeins}
\usepackage{float}
\usepackage[percent]{overpic}

\definecolor{cvprblue}{rgb}{0.21,0.49,0.74}
\definecolor{resultshade}{RGB}{239,245,248}
\definecolor{resultfocus}{RGB}{219,233,242}
\usepackage[pagebackref,breaklinks,colorlinks,allcolors=cvprblue]{hyperref}
\def\confName{CVPR}
\def\confYear{2027}

\title{EgoEMG: A Multimodal Egocentric Dataset with Bilateral EMG and Vision for Hand Pose Estimation}
\author{%
Ziheng Xi$^{1}$, Jiayi Yu$^{1}$, Yanbo Duan$^{1}$, Yitao Wang$^{1}$,\\
Jinyang Yu$^{2}$, Zhanwei Xu$^{2}$, Jianjiang Feng$^{1}$, Jie Zhou$^{1}$\\
$^{1}$\href{https://www.tsinghua.edu.cn}{Tsinghua University} \quad $^{2}$\href{https://showee.top}{Dexwise}\\
{\scriptsize\ttfamily \{xizh21, jy-yu23, duanyb24, wangyita23\}@mails.tsinghua.edu.cn  }\\
{\scriptsize\ttfamily \{yujinyang, xuzhanwei\}@showee.top \quad \{jfeng, jzhou\}@tsinghua.edu.cn}
}
\date{}

\newcommand{\dataset}{\textit{EgoEMG}}
\newcommand{\model}{EMGFormer}
\newcommand{\marker}{\textit{markers2mano}}

\newcommand{\supplementref}[1]{Appendix~\ref{#1}}
\newcommand{\mainref}[1]{\ref{#1}}

\begin{document}

\maketitle

\begin{abstract}
Hand pose estimation from egocentric video is challenging under adverse visual conditions, such as occlusion. Wrist-worn surface electromyography (sEMG) measures muscle activity without requiring visual access to the hand, offering a wearable complement to egocentric cameras. Evaluating their combined use requires synchronized physiological and visual recordings with continuous pose annotations, which existing EMG-based hand pose datasets do not jointly provide. We introduce EgoEMG, a multimodal dataset pairing bilateral wrist sEMG with egocentric video and motion-capture-derived hand poses, supplemented by external multi-view RGB-D recordings. EgoEMG contains 25.9 hours of recordings from 68 participants, covering 60 bare-hand gesture classes and interactions with 10 everyday objects. We establish EMG-only, vision-only, and fusion benchmarks with shared 22-dimensional pose targets and evaluations on unseen users, gestures, and objects. For EMG-only estimation, EMGFormer achieves the lowest overall MAE among the compared methods on EgoEMG and competitive results on EMG2Pose. Our residual fusion model combines ViT and EMGFormer features. Adding EMG reduces error for both single-frame and temporal vision models, with MAE decreasing from $7.39^\circ$ to $7.06^\circ$ for ViT-L. The largest absolute reduction occurs when evaluating interactions with an unseen object. Generalization to unseen users remains challenging.
\par\noindent Project page: \url{https://zhenqis123.github.io/egoemg-site/}.
\end{abstract}

\section{Introduction}
Hand pose estimation supports augmented reality, human--computer interaction,
and embodied AI. Egocentric cameras enable hand tracking~\citep{hamer,h2o},
but occlusion, limited field of view, motion blur, and poor lighting limit
pose estimates during hand movements.
These limitations motivate complementary wearable sensing that does not
depend on hand visibility.

Wrist-worn surface EMG directly measures muscle activity~\citep{farina2014}.
Recent methods recover continuous hand articulation from these
signals~\citep{emg2pose,emg2tendon,vqmyopose}, suggesting that EMG could supplement
egocentric vision when visual evidence is incomplete. Evaluating this
possibility requires synchronized recordings with hand pose annotations.
Existing EMG pose datasets lack egocentric video, limiting direct comparisons
of EMG, vision, and their fusion.

We introduce \dataset{}, a multimodal dataset for bimanual hand pose
estimation. It pairs bilateral wrist EMG and egocentric video with
motion-capture-derived joint angles, supplemented by IMU and
external multi-view RGB-D recordings. The dataset contains 25.9 hours of
recordings from 68 participants, covering 60 bare-hand gesture classes and
interactions with 10 everyday objects. Continuous pose labels for both hands
support evaluation of finger and wrist articulation throughout these
activities. We also provide object meshes reconstructed from laser scans,
segmentation masks, and object pose estimates to support multimodal studies
of hand--object interaction (HOI).

We benchmark EMG-only, vision-only, and fused estimation under shared
evaluation conditions for unseen users, unseen gestures, their
intersection, and unseen objects.
\model{} achieves the lowest overall MAE among the compared methods on
\dataset{} and competitive results on EMG2Pose. Experiments with six visual
backbones and residual fusion show that adding EMG reduces joint-angle
MAE for both single-frame and temporal vision models on all the splits. For single-frame ViT-L,
fusion reduces MAE from $7.39^\circ$ to $7.06^\circ$. The largest error reductions occur in the
Object split. Generalization to unseen users remains challenging.

Our contributions are:
\begin{itemize}[leftmargin=1.5em,nosep]
    \item We introduce \dataset{}, a multimodal dataset linking
    bilateral wrist EMG, egocentric video, and continuous hand
    pose labels across bare-hand movements and hand-object interactions.
    \item We establish benchmarks for EMG, vision, and their fusion under
    shared pose targets and generalization splits, and introduce \model{}
    for continuous EMG-based hand pose estimation.
    \item We develop a residual EMG--vision fusion model and show that adding EMG
    reduces joint-angle MAE across tested vision-based models.
\end{itemize}

\begin{figure*}[t]
\centering
\begin{overpic}[width=\textwidth]{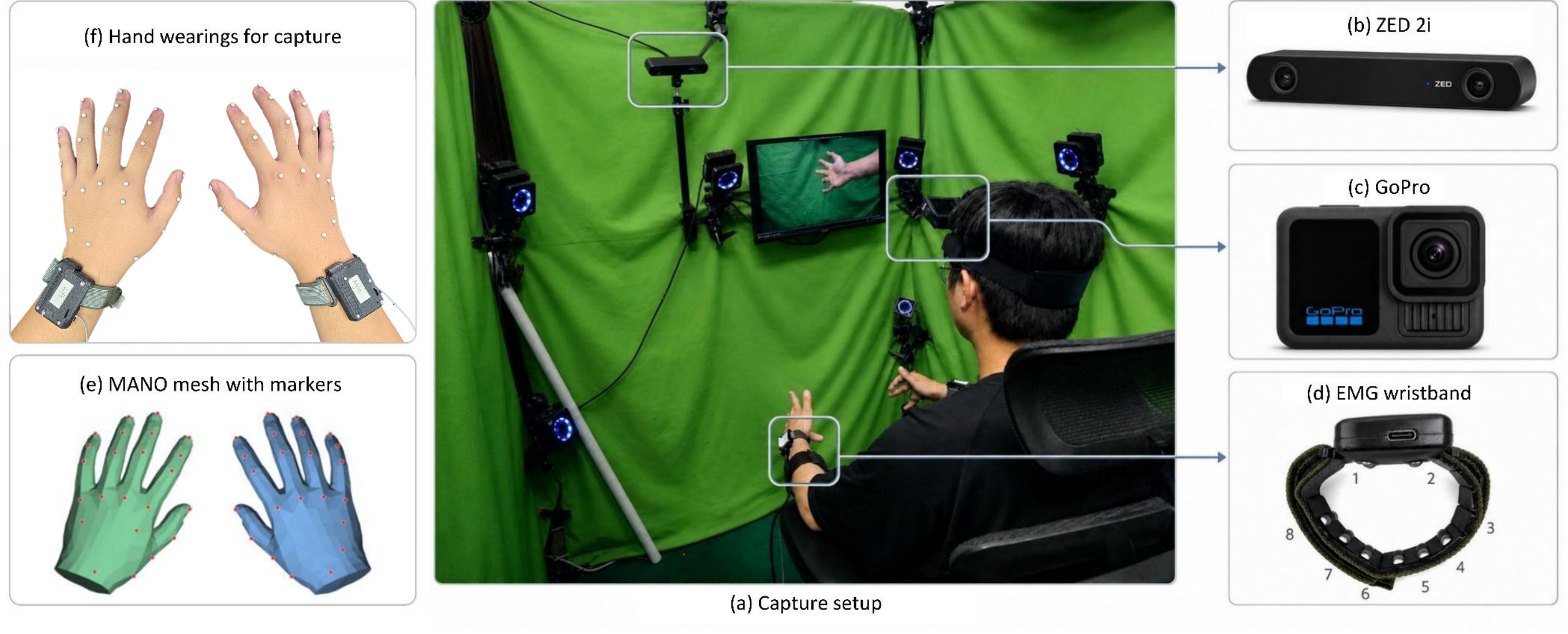}
\put(3.2,36.9){\colorbox{white}{\parbox[c][10pt][c]{0.210\textwidth}{\centering\sffamily\fontsize{6}{7}\selectfont (f) Hand markers and EMG wristbands}}}
\end{overpic}
\caption{Data collection setup for \dataset{}. We record bilateral wrist EMG, head-mounted egocentric RGB, and external ZED 2i RGB-D, and obtain pose labels from optical motion capture with hand markers.}
\label{fig:capture_setup}
\end{figure*}

\section{Related Work}

\paragraph{Hand pose datasets and benchmarks.}
EMG2Pose~\citep{emg2pose} pairs wrist EMG with
motion-capture-derived joint angles, while MyoKi~\citep{myoki} and
\citet{jarque2019} record forearm EMG with instrumented-glove kinematics.
Other EMG datasets emphasize gesture recognition~\citep{ninapro,grabmyo,putemg}.
Visual datasets cover monocular hand reconstruction~\citep{freihand},
interacting hands~\citep{interhand}, and hand--object pose~\citep{dexycb}.
H2O~\citep{h2o} and HOT3D~\citep{hot3d} provide egocentric recordings,
whereas GigaHands~\citep{gigahand} captures bimanual activities with a fixed
multi-camera system. \dataset{} brings egocentric video and bilateral EMG
together with continuous pose labels for comparison on shared targets.
Table~\ref{tab:dataset_compare} summarizes the sensing and annotation coverage.

\paragraph{Vision-based hand pose estimation.}
The MANO model~\citep{mano} provides a parametric representation of hand
shape and articulation. HaMeR~\citep{hamer} and WiLoR~\citep{wilor} use
Transformers for hand reconstruction from monocular RGB images.
UmeTrack~\citep{umetrack} addresses multi-view hand tracking for VR,
while WristPP~\citep{xi2026wristpp} estimates hand pose and pressure from
a wrist-mounted camera.

\paragraph{EMG-based hand pose estimation.}
Early work estimated continuous finger kinematics from muscle-activation
models~\citep{ngeo2012continuous,ngeo2014continuous};
NeuroPose~\citep{liu2021neuropose} decodes 3D hand pose from wearable EMG. Recent methods explore tendon control~\citep{emg2tendon},
quantized pose representations~\citep{vqmyopose}, user
adaptation~\citep{react2026}, and dexterous teleoperation~\citep{zhao2026dexemg}.
\citet{hadidi2026reevaluating} compare position and velocity decoding on
EMG2Pose under a causal evaluation protocol.
TinyMyo~\citep{fasulo2025tinymyo} studies compact EMG pretraining across
multiple tasks, including regression of hand kinematics.

\paragraph{Cross-modal learning and fusion.}
EMG has been fused with image-derived hand skeletons for gesture
recognition~\citep{hao2024multimodal} and with gaze-guided egocentric
vision for grasp-intent inference~\citep{zandigohar2024multimodal}.
CPEP~\citep{cpep} and EMBridge~\citep{embridge} align EMG with pose
representations to improve gesture generalization.
\citet{liu2025frompose} predict EMG from video-derived hand poses
and piano keystrokes. PIMForce~\citep{pimforce} uses EMG and posture to estimate
hand pressure, while ForceBand~\citep{forceband} combines force estimates
from EMG and IMU with egocentric video for robot learning.
Our benchmark compares continuous hand-pose estimation from EMG, vision,
and their fusion under common labels and evaluation conditions.

\section{\dataset{} Dataset}
\label{sec:dataset}

\begin{figure*}[t]
\centering
\includegraphics[width=\textwidth]{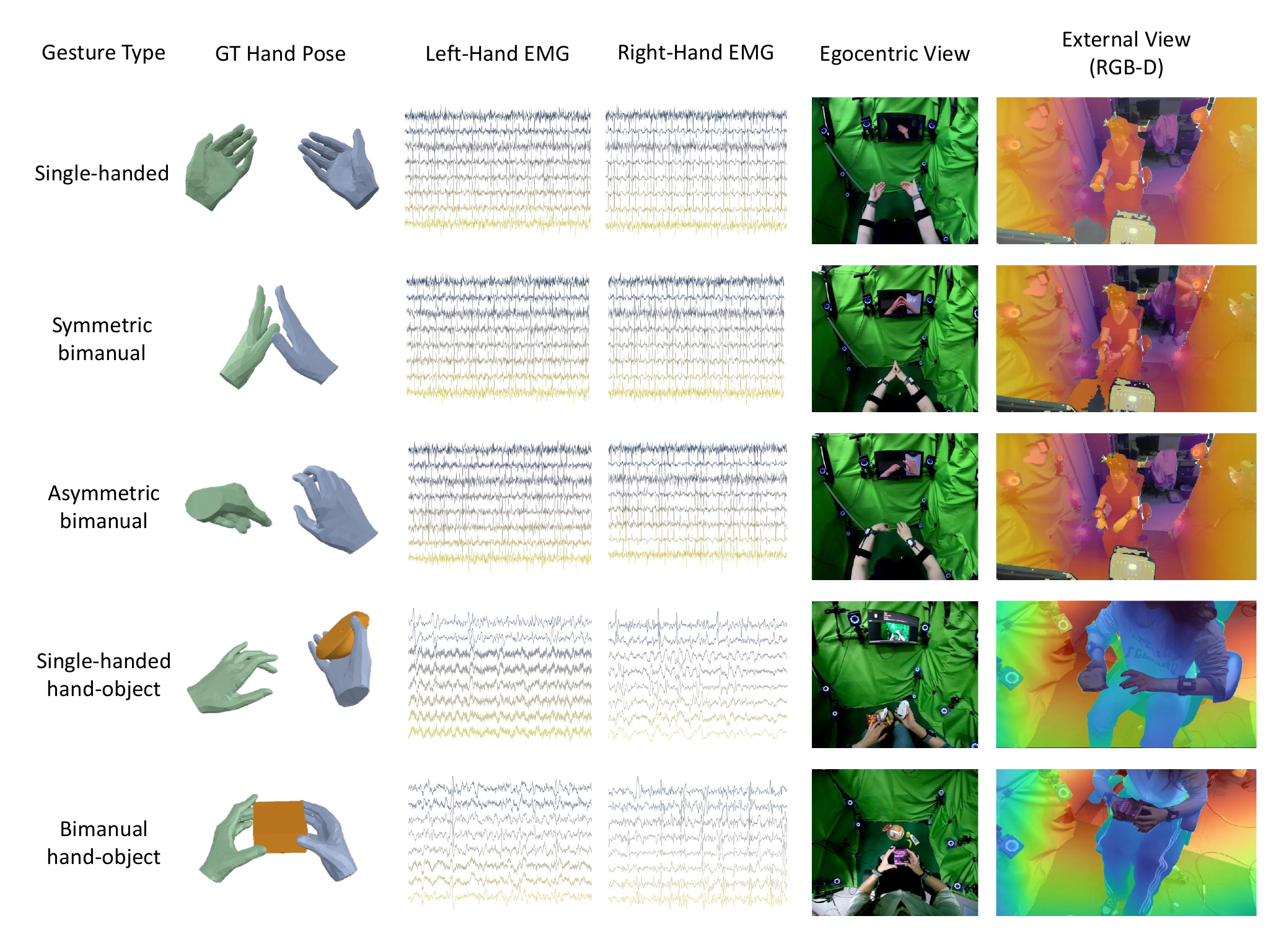}
\caption{Synchronized bare-hand and hand--object interaction samples from \dataset{}. Each row shows 6\,s of bilateral EMG, with hand poses, egocentric RGB, and external RGB-D at the window center.}
\label{fig:sample_data}
\end{figure*}

\subsection{Overview and Capture Setup}

\dataset{} records bilateral EMG and IMU, egocentric RGB, external RGB-D, and optical motion capture. Each wrist carries a WAVELETECH wristband that records 8 EMG channels at 2\,kHz.
A GoPro 13 camera records egocentric RGB during bare-hand sessions, while a
Dexwise head-mounted camera\footnote{\url{https://www.showee.top/}} is used for hand-object sessions. An external ZED RGB-D camera
records all sessions at 30\,Hz. Two additional RealSense D455 cameras record
HOI sessions to support object pose estimation.
Hand motion is tracked using 21 reflective markers per hand at 120\,Hz with an FZMotion optical motion capture system\footnote{\url{https://www.ymate3d.com/}}. All modalities share a host-clock time base and are aligned using timestamps
(Figure~\ref{fig:capture_setup}). Alignment rules and residuals are reported in
\supplementref{app:dataset_card}.

Recordings cover bare-hand gestures, hand--hand coordination, and
hand--object interaction. The gesture vocabulary contains 60 classes
(30 single-hand, 30 bimanual) with per-frame labels for evaluation by gesture. Gesture prompts appear in random order. For the single-hand gesture classes, participants perform the same gesture
symmetrically with both hands while varying wrist orientation. For the
bimanual classes, they follow demonstrations. The full vocabulary is listed in
\supplementref{app:gesture_vocabulary}. The HOI recordings cover interactions with 10 everyday objects. Figure~\ref{fig:sample_data} shows synchronized examples of bare-hand motion and hand--object interaction.

\subsection{Pose Label Reconstruction}
\label{subsec:label_reconstruction}

We reconstruct MANO pose and shape parameters~\citep{mano} from the
21 markers on each hand using \marker{}, with fixed surface vertices
defining marker-to-mesh correspondences. The reconstructed meshes achieve
a mean marker-to-mesh alignment error of 4.3\,mm.
The reconstruction procedure and estimated hand-shape variation are
detailed in \supplementref{app:markers2mano} and
\supplementref{app:mano_shape_diversity}, respectively.

To obtain the 22-dimensional pose targets, we fit the UmeTrack hand model
directly to the observed 3D markers using fixed surface-marker attachments,
yielding 20 finger joint angles per hand. The mean per-frame RMS
marker-fitting residual is 2.8\,mm
(\supplementref{app:joint_angle_conversion}).
We compute two armband-referenced wrist angles for flexion/extension and
radial/ulnar deviation from motion-capture markers, with reference directions
and zero angles defined by the armband marker geometry
(\supplementref{app:wrist_angles}).
The resulting 22-dimensional vectors use the same output ordering for both hands.

Across bare-hand and HOI recordings, the reconstruction pipeline marks
3.6\% of frames as invalid. For qualitative inspection, we reproject the
reconstructed meshes onto egocentric images using camera calibration.
Figures~\ref{fig:label_quality} and~\ref{fig:hoi_label_quality}
show gesture and object interaction examples, respectively.
\subsection{Dataset Statistics and Positioning}
\label{sec:data_partitioning}

\dataset{} comprises 81 recording sessions from 68 participants,
totaling 25.92\,hours: 11.80\,hours of bare-hand motion and
14.13\,hours of hand--object interaction (HOI).
EMG, egocentric RGB, and 22-dimensional pose labels are jointly available
for 89.71\% of the recorded data.
\supplementref{app:dataset_card} reports coverage by activity and split
(Table~\ref*{tab:modality_coverage}), duration accounting, and validity criteria.

The annotated HOI portion covers 16 participants interacting with
10 everyday objects. Annotations include object identities, task
intervals, object meshes, segmentation masks, and 6D object poses.
Segmentation masks and object poses are generated using SAM~3~\citep{sam3}
and FoundationPose~\citep{foundationposewen2024}, respectively. Annotation coverage varies across
recordings, as detailed in \supplementref{app:hoi_coverage}.

Table~\ref{tab:dataset_compare} compares \dataset{} with representative
public EMG datasets. Among those compared, EMG2Pose and \dataset{}
provide bilateral EMG with motion-capture-derived hand-pose labels.
\dataset{} additionally provides synchronized egocentric video,
external RGB-D recordings, wrist articulation angles, and object-level
annotations, supporting multimodal hand pose estimation during both
bare-hand motion and object interaction.

\begin{table*}[t]
\centering
\caption{Comparison of \dataset{} with representative public EMG datasets.
``EMG Ch.'' counts channels per recorded arm (per wristband for bilateral datasets); ``Bilateral'' indicates simultaneous EMG capture from both arms; ``Wrist labels'' indicates explicit wrist articulation angles; ``Pose labels'' gives the number of joint angles unless noted; ``Object annotations'' indicates explicit object-level labels, such as object identities, poses, or segmentation masks, beyond action or gesture labels.}
\label{tab:dataset_compare}
\footnotesize
\setlength{\tabcolsep}{3pt}
\renewcommand{\arraystretch}{1.12}
\begin{threeparttable}
\begin{tabular*}{\textwidth}{@{\extracolsep{\fill}}lccccccccc@{}}
\toprule
Dataset & Subj. & EMG Ch. & Bilateral & Ego RGB & RGB-D &
\makecell{Wrist\\labels} & \makecell{Pose\\labels} & \makecell{Gestures/\\tasks} & \makecell{Object\\annotations} \\
\midrule
Jarque-Bou et al.~\citep{jarque2019} & 22 & 7 & -- & -- & -- & \checkmark & 18 & 26 & -- \\
NinaPro DB1~\citep{ninapro} & 27 & 10 & -- & -- & -- & -- & 22\tnote{*} & 53 & -- \\
MyoKi~\citep{myoki} & 35 & 12 & -- & -- & -- & \checkmark & 18 & 74 & -- \\
PIMForce~\citep{pimforce} & 21 & 8 & -- & -- & -- & -- & 20 & 22 & -- \\
EMG2Pose~\citep{emg2pose} & 193 & 16 & \checkmark & -- & -- & -- & 20 & 50 & -- \\
\textbf{\dataset{} (ours)}
& \textbf{68} & \textbf{8} & \checkmark & \checkmark & \checkmark
& \checkmark & \textbf{22} & \textbf{60} & \checkmark \\
\bottomrule
\end{tabular*}
\begin{tablenotes}
\item[*] NinaPro DB1 provides 22 uncalibrated glove-sensor channels; its 53 classes include rest. MyoKi records across the forearm and upper arm. Activity counts reflect each dataset's own gesture or task definitions.
\item[] The object-annotated HOI portion of \dataset{} covers 10 objects, separate
from the 60 gesture classes, with object-task identities and automatic
segmentation masks. Modality validity varies by sample.
\end{tablenotes}
\end{threeparttable}
\end{table*}

\paragraph{Data partitions.}

The training, validation, and test sets contain 20.73, 2.62, and
2.57\,hours, respectively. Following EMG2Pose~\citep{emg2pose}, we evaluate
unseen gestures (Gesture), unseen participants (User), and their
intersection (G+U); Table~\ref{tab:evaluation_splits} summarizes these
conditions. User and G+U use two held-out participants for validation
and three for testing. The held-out gestures are listed in
\supplementref{app:split_details}.

The Object condition tests unseen objects with participants seen in
training. The bleach cleanser bottle is held out for validation and the
pudding mix box for testing; the other eight objects provide training
examples.

\begin{table}[t]
\centering
\caption{Generalization conditions used for validation and testing.
Seen and unseen refer to the training set; dashes indicate inapplicable axes.}
\label{tab:evaluation_splits}
\footnotesize
\setlength{\tabcolsep}{4pt}
\renewcommand{\arraystretch}{1.12}
\begin{tabular*}{\columnwidth}{@{\extracolsep{\fill}}lccc@{}}
\toprule
Condition & Participants & Gestures & Objects \\
\midrule
Gesture & Seen & Unseen & -- \\
User & Unseen & Seen & -- \\
G+U & Unseen & Unseen & -- \\
Object & Seen & -- & Unseen \\
\bottomrule
\end{tabular*}
\end{table}

All modalities use the same split, with session lists provided in the
dataset release. EMG windows and RGB sequences do not cross data splits,
recording gaps, or task boundaries. Models are selected using validation
MAE over all four conditions; test data are reserved for final reporting.

\section{Benchmark and Baselines}

\subsection{Benchmark Protocol}
\label{sec:benchmark_protocol}

We evaluate EMG-to-pose, vision-to-pose, and EMG+vision fusion using the
same joint-angle targets and data partitions
(Section~\ref{sec:data_partitioning}). EMG-only models predict joint angles
throughout four-second EMG windows. Fusion uses the same windows and predicts
joint angles at the center frame. Vision models take the resized
$256\times256$ egocentric image as input.

We report mean absolute error (MAE) in degrees over valid joint angles.
Let $\hat y_{htj}$ and $y_{htj}$ be predicted and target angles in radians
for hand $h$, time $t$, and joint $j$, with validity mask $m_{htj}$.
For an evaluation set $\mathcal E$,
\begin{equation}
\operatorname{MAE}_{\mathcal E}
=\frac{180}{\pi}\,
\frac{\sum_{(h,t)\in\mathcal E}\sum_{j=0}^{21}
m_{htj}|\hat y_{htj}-y_{htj}|}
{\sum_{(h,t)\in\mathcal E}\sum_{j=0}^{21}m_{htj}}.
\label{eq:benchmark_mae}
\end{equation}
Missing pose labels are excluded. Overall MAE is computed over all valid
joint angles across both hands and the four evaluation conditions. The training loss uses the
same valid angles and is computed in radians (\supplementref{app:training_objective}).

\supplementref{app:emg_center_test} additionally reports center-frame MAE
for EMG-only models.

\subsection{EMG Baselines}

\begin{figure}[t]
    \centering
    \includegraphics[width=0.9\columnwidth]{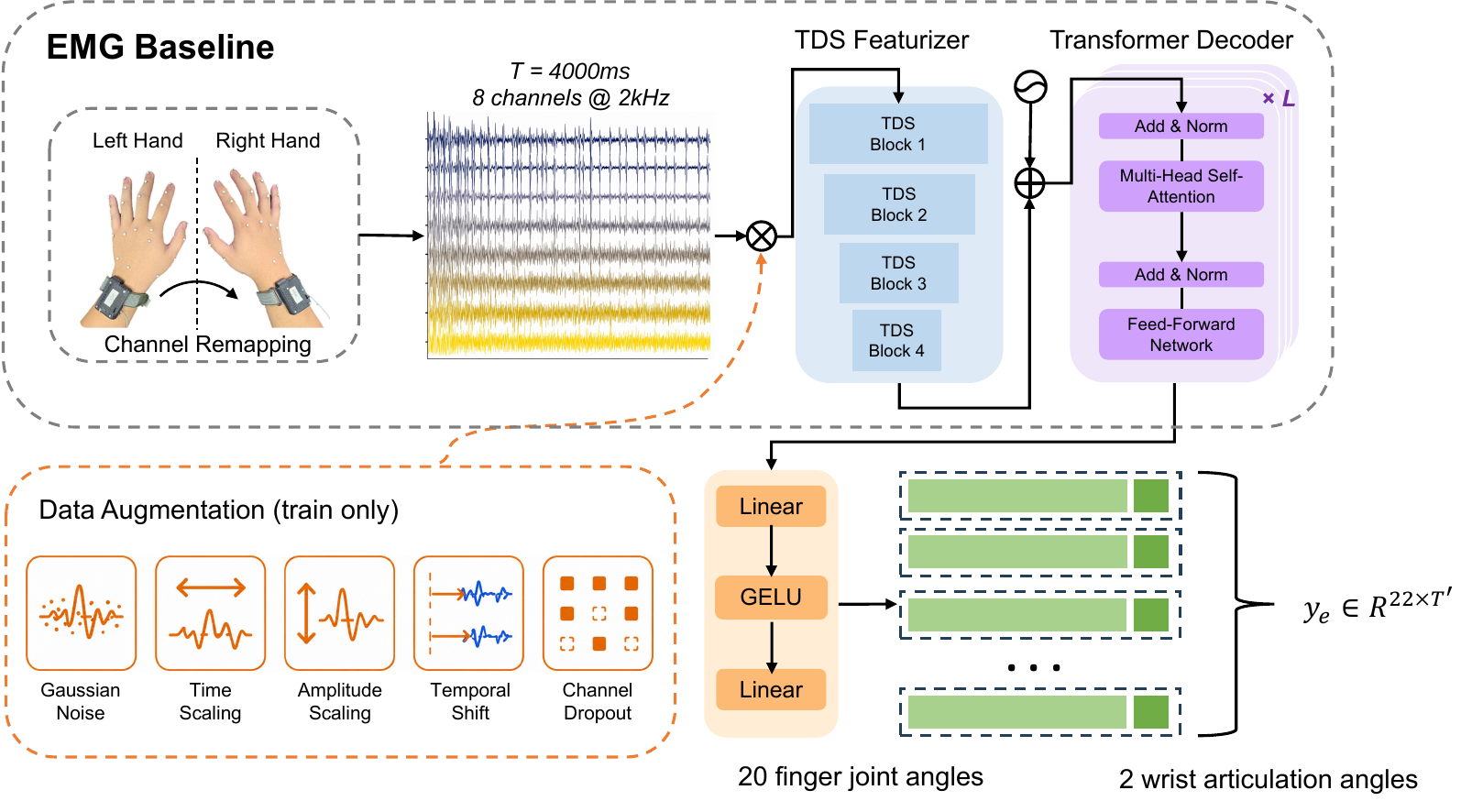}
    \caption{\model{} architecture for EMG-to-pose. A TDS-style convolutional
    encoder and a Transformer with RoPE process EMG from one wrist. A regression
    head predicts finger and wrist joint angles over time.}
    \label{fig:architecture}
\end{figure}

We compare \model{} with emg2pose and vemg2pose~\citep{emg2pose},
NeuroPose~\citep{liu2021neuropose}, and
SensingDynamics~\citep{sensingdynamics}. \model{} takes four seconds of
filtered, eight-channel EMG from one wrist and predicts that hand's joint
angles over time (Figure~\ref{fig:architecture}). A TDS-style convolutional
encoder~\citep{tds} extracts features and reduces the sequence length.
A Transformer with rotary positional encoding (RoPE)~\citep{rope} captures
temporal dependencies, followed by a regression head that predicts joint angles.
During dataset preparation, left-wrist EMG channels are reordered to
match anatomically corresponding channels on the right wrist, accounting
for the mirrored anatomy of the two hands. We evaluate Small, Medium, and Large variants; architecture
and training settings are given in \supplementref{app:training_details}.

\subsection{Vision Baselines}

\begin{figure}[t]
    \centering
    \includegraphics[width=\columnwidth,pagebox=cropbox]{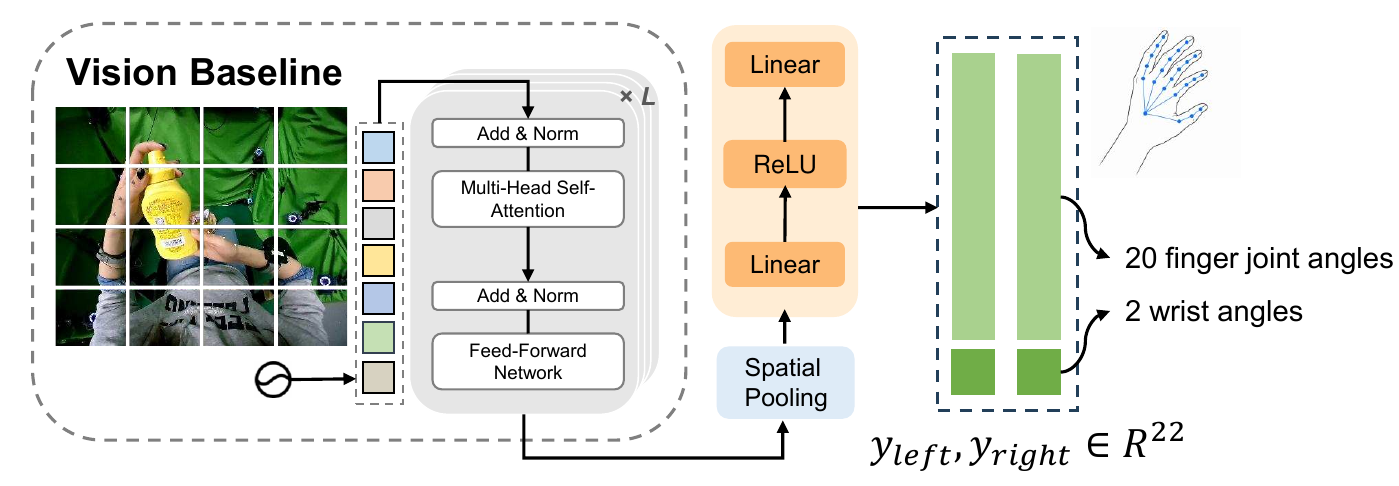}
    \caption{Vision-to-pose baseline. An image encoder and a regression head
    predict both hands' joint angles from the entire egocentric image.
    The diagram shows the ViT variant.}
    \label{fig:vision_architecture}
\end{figure}

We evaluate ResNet-\{18, 50, 152\}~\citep{resnet} and
ViT-\{Small, Base, Large\}~\citep{vit}. ResNet models use
ImageNet-pretrained weights~\citep{imagenet}, while ViT models use
DINOv2-pretrained weights~\citep{dinov2}.
An image encoder extracts a feature vector $\mathbf z_t$ from the entire
image $I_t$. An MLP predicts the joint angles of both hands,
$[\hat{\mathbf y}_{v,t}^{L};\hat{\mathbf y}_{v,t}^{R}]$ (Figure~\ref{fig:vision_architecture}). Head dimensions and
optimization settings are provided in
\supplementref{app:training_details_vision}.

\paragraph{Temporal visual baseline.}
To test whether EMG remains useful when the vision model can also use
neighboring frames, we evaluate a temporal ViT-S baseline. It processes
17 images at 0.25\,s intervals over
$[t-2\,\mathrm{s},t+2\,\mathrm{s}]$, covering the same four seconds as the
EMG window. A Transformer combines information across these images to
update the center-frame feature, which is then used to predict both hands'
joint angles. Temporal fusion uses this feature in the residual fusion
model described next. The temporal architecture is detailed in
\supplementref{app:temporal_visual_control}.

\subsection{Residual Fusion}
\label{sec:fusion_arch}

\begin{figure}[t]
    \centering
    \includegraphics[width=\columnwidth,pagebox=cropbox]{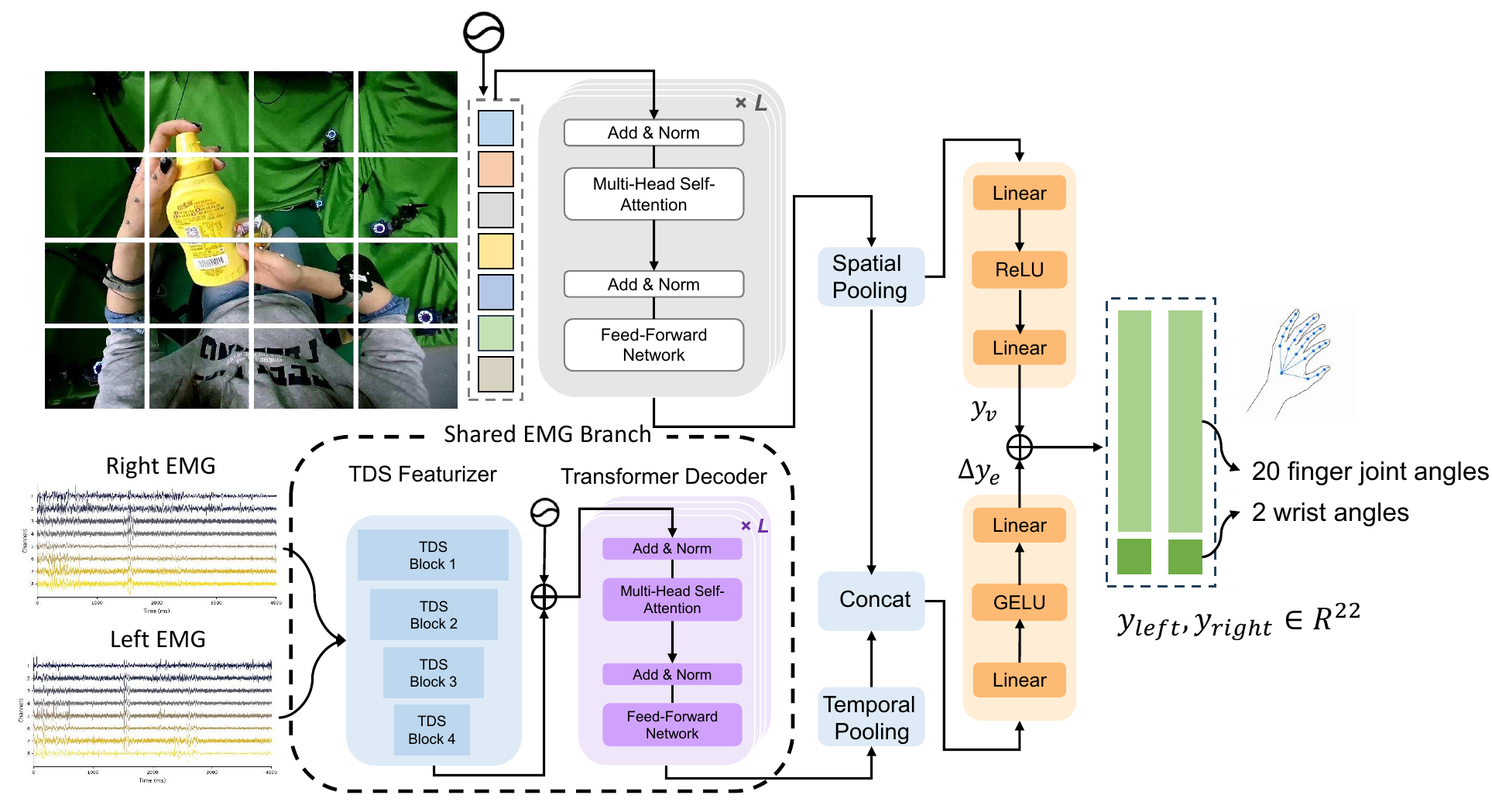}
    \caption{EMG+vision fusion baseline. The image encoder processes the entire
    egocentric image. Both hands share the EMG encoder and temporal pooling,
    while separate residual heads refine their joint-angle predictions
    (Eq.~\ref{eq:bimanual_residual_fusion}). Layer details are provided in
    \supplementref{app:training_details_vision}.}
    \label{fig:fusion_architecture}
\end{figure}

The fusion model uses EMG to refine the joint angles estimated from the
image for each hand (Figure~\ref{fig:fusion_architecture}). The visual
branch and \model{}-S branch are initialized from pretrained vision-only
and EMG-only models, respectively, and fine-tuned jointly.
The shared EMG encoder processes each wrist independently, and temporal
attention pooling combines its features over time into a vector
$\mathbf e_t^h$ for hand $h$. We concatenate this vector with the projected
image feature $P_v(\mathbf z_t)$ and an encoding $\mathbf q_h$ that
identifies the left or right hand:
\begin{equation}
\begin{aligned}
\mathbf{u}_t^h &= G([\mathbf{e}_t^h;P_v(\mathbf{z}_t);\mathbf{q}_h]),\\
\hat{\mathbf{y}}_t^h &= \hat{\mathbf{y}}_{v,t}^h + H_h(\mathbf{u}_t^h).
\end{aligned}
\label{eq:bimanual_residual_fusion}
\end{equation}
The projection $G$ is shared across hands, while separate heads $H_L$
and $H_R$ predict residuals. The residual heads are initialized to produce
near-zero corrections, preserving
the initial visual predictions. The loss is applied only to the final
joint-angle predictions at the center frame.
Layer dimensions are given in
\supplementref{app:training_details_vision}.

\section{Experimental Results}

\subsection{EMG Baseline Results}

\model{}-S achieves the lowest Overall MAE on \dataset{}
(Table~\ref{tab:egoemg_emg_results}), with $12.28^\circ$ compared with
$12.76^\circ$ for emg2pose (a $3.8\%$ reduction). Increasing model size does
not reduce Overall MAE: both M and L achieve $12.46^\circ$. \model{}-L achieves the lowest MAE
under the User and G+U conditions, while \model{}-S achieves the lowest MAE
under the Object condition, at $9.92^\circ$. We use \model{}-S, with
3.5M parameters, as the EMG backbone for fusion.
Causal variants yield Overall MAE within $0.17^\circ$ of the bidirectional
models, while increasing error under the User condition for all three
sizes (\supplementref{app:emg_model_causal}). These variants use causally
filtered EMG and restrict inference to current and past inputs,
making them suitable for real-time inference.

\begin{table}[t]
\centering
\caption{EMG-to-pose MAE ($^\circ$, $\downarrow$). Shading marks the compact S baseline; bold marks the lowest MAE in each column. Overall pools valid angles across all four conditions.}
\label{tab:egoemg_emg_results}
\footnotesize
\setlength{\tabcolsep}{3pt}
\renewcommand{\arraystretch}{1.12}
\resizebox{\columnwidth}{!}{%
\begin{tabular}{lrrrrrr}
\toprule
Method & Params & Gesture & User & G+U & Object & Overall \\
\midrule
\rowcolor{resultshade}
\model{}-S & 3.5M & $\mathbf{13.40}$ & 15.37 & 16.87 & $\mathbf{9.92}$ & \cellcolor{resultfocus}$\mathbf{12.28}$ \\
\model{}-M & 6.6M & $\mathbf{13.40}$ & 15.69 & 17.19 & 10.15 & 12.46 \\
\model{}-L & 16.3M & 13.92 & $\mathbf{14.83}$ & $\mathbf{16.14}$ & 10.29 & 12.46 \\
\midrule
NeuroPose & 6.4M & 16.25 & 15.79 & 16.62 & 12.63 & 14.45 \\
emg2pose & 3.0M & 14.19 & 16.06 & 17.04 & 10.16 & 12.76 \\
SensingDynamics & 1.0M & 15.82 & 16.11 & 16.24 & 11.39 & 13.43 \\
vemg2pose & 6.0M & 14.14 & 16.67 & 17.69 & 10.37 & 13.00 \\
\bottomrule
\end{tabular}}%
\end{table}

\paragraph{Results on EMG2Pose.}
Table~\ref{tab:emg2pose_results} reports results on EMG2Pose~\citep{emg2pose};
evaluation settings are summarized in \supplementref{app:emg2pose_protocol}. The User, Stage, and User+Stage conditions evaluate generalization
to unseen participants, unseen movement categories, and both, respectively.
\model{}-L achieves the lowest User and User+Stage MAEs in the table,
$11.2^\circ$ and $13.9^\circ$, respectively, compared with the reported
$11.3^\circ$ for CLDM on User and $14.6^\circ$ for Position and
Velocity MT on User+Stage~\citep{hadidi2026reevaluating}.
\model{}-S achieves the lowest Stage MAE, $12.9^\circ$, compared with
$13.9^\circ$ for Velocity MT. Scaling from S to L improves User and
User+Stage MAE but increases Stage MAE from $12.9^\circ$ to $13.2^\circ$.

\begin{table}[t]
\centering
\caption{EMG-only MAE ($^\circ$, $\downarrow$) on EMG2Pose; mean $\pm$ SD across users. $^\dagger$: reported results; MT: joint Regression+Tracking training, evaluated in Regression mode. NeuroPose and SensingDynamics MAEs are from the EMG2Pose benchmark~\citep{emg2pose}. Evaluation protocols are summarized in \supplementref{app:emg2pose_protocol}. Bold marks the lowest mean per column at the displayed precision.}
\label{tab:emg2pose_results}
\footnotesize
\setlength{\tabcolsep}{3.5pt}
\renewcommand{\arraystretch}{1.05}
\resizebox{\columnwidth}{!}{%
\begin{tabular}{lcccc}
\toprule
Method & Params & User & Stage & User+Stage \\
\midrule
\multicolumn{5}{l}{\textit{\model{} variants}} \\
\midrule
\model{}-S & 3.5M & $11.4_{\pm1.1}$ & $\mathbf{12.9_{\pm1.5}}$ & $14.1_{\pm1.2}$ \\
\model{}-M & 6.6M & $11.3_{\pm1.0}$ & $13.6_{\pm1.6}$ & $14.1_{\pm1.1}$ \\
\rowcolor{resultshade}
\model{}-L & 16.3M & $\mathbf{11.2_{\pm1.1}}$ & $13.2_{\pm1.6}$ & $\mathbf{13.9_{\pm1.2}}$ \\
\midrule
\multicolumn{5}{l}{\textit{Prior work (reported)}} \\
\midrule
CLDM$^\dagger$~\citep{emg2tendon} & 7.0M & $11.3_{\pm1.0}$ & $14.3_{\pm1.5}$ & $14.7_{\pm1.4}$ \\
Position MT$^\dagger$~\citep{hadidi2026reevaluating} & 6.0M & $11.5_{\pm1.2}$ & $14.0_{\pm1.6}$ & $14.6_{\pm1.3}$ \\
Velocity MT$^\dagger$~\citep{hadidi2026reevaluating} & 6.0M & $11.6_{\pm1.3}$ & $13.9_{\pm1.6}$ & $14.6_{\pm1.3}$ \\
vemg2pose$^\dagger$~\citep{emg2pose} & 6.0M & $12.2_{\pm1.3}$ & $15.2_{\pm1.6}$ & $15.8_{\pm1.4}$ \\
emg2pose (position)$^\dagger$~\citep{emg2pose} & 3.0M & $12.6_{\pm1.3}$ & $15.2_{\pm1.6}$ & $15.6_{\pm1.3}$ \\
NeuroPose$^\dagger$~\citep{liu2021neuropose} & 6.4M & $13.2_{\pm1.1}$ & $17.2_{\pm1.7}$ & $17.5_{\pm1.5}$ \\
SensingDynamics$^\dagger$~\citep{sensingdynamics} & 1.0M & $15.5_{\pm1.4}$ & $18.8_{\pm1.6}$ & $18.7_{\pm1.6}$ \\
\bottomrule
\end{tabular}}%
\end{table}

\subsection{Vision and Fusion Results}
\label{sec:cross_modal}

Larger ViT backbones improve vision-only estimation
(Table~\ref{tab:egoemg_results}): Overall MAE decreases from $8.55^\circ$
for ViT-S/14 to $8.02^\circ$ for ViT-B/14 and $7.39^\circ$ for ViT-L/14.
All three backbones have higher errors under User and G+U than under Gesture.

When evaluated on the same center frames, fusion outperforms vision alone
in all four conditions for each ViT backbone. Reductions in Overall MAE
range from $0.33^\circ$ to $0.51^\circ$, with the largest reduction for
ViT-S/14. Fusion with ViT-L/14 achieves the lowest Overall MAE, $7.06^\circ$.
The largest absolute reduction occurs under the Object condition for each
backbone, ranging from $0.40^\circ$ to $0.68^\circ$.
ResNet results are provided in
\supplementref{app:resnet_results}.

\begin{table}[t]
\centering
\caption{ViT vision and EMG+vision MAE ($^\circ$, $\downarrow$). Shaded rows add EMG. Fusion gain is vision MAE minus fusion MAE, computed before rounding. Bold MAE values mark the lowest error in each column; darker blue cells highlight Object fusion gains.}
\label{tab:egoemg_results}
\footnotesize
\setlength{\tabcolsep}{1.5pt}
\renewcommand{\arraystretch}{1.12}
\begin{tabular*}{\columnwidth}{@{\extracolsep{\fill}}llrrrrr@{}}
\toprule
Backbone & & Gesture & User & G+U & Object & Overall \\
\midrule
\multirow[t]{3}{*}{ViT-S/14} & Vision & 7.68 & 8.74 & 9.09 & 9.02 & 8.55 \\
\rowcolor{resultshade}
\cellcolor{white} & + EMG & 7.36 & 8.33 & 8.47 & 8.34 & 8.05 \\
 & Fusion gain & 0.32 & 0.41 & 0.62 & \cellcolor{resultfocus}0.68 & 0.51 \\
\addlinespace[2pt]
\multirow[t]{3}{*}{ViT-B/14} & Vision & 7.15 & 8.13 & 8.33 & 8.55 & 8.02 \\
\rowcolor{resultshade}
\cellcolor{white} & + EMG & 6.87 & 7.85 & 7.96 & 8.07 & 7.65 \\
 & Fusion gain & 0.28 & 0.28 & 0.37 & \cellcolor{resultfocus}0.48 & 0.37 \\
\addlinespace[2pt]
\multirow[t]{3}{*}{ViT-L/14} & Vision & 6.55 & 7.69 & 7.67 & 7.81 & 7.39 \\
\rowcolor{resultshade}
\cellcolor{white} & + EMG & $\mathbf{6.23}$ & $\mathbf{7.48}$ & $\mathbf{7.32}$ & $\mathbf{7.41}$ & $\mathbf{7.06}$ \\
 & Fusion gain & 0.32 & 0.21 & 0.35 & \cellcolor{resultfocus}0.40 & 0.33 \\
\bottomrule
\end{tabular*}
\end{table}

\begin{figure*}[t]
\centering
\includegraphics[width=\textwidth]{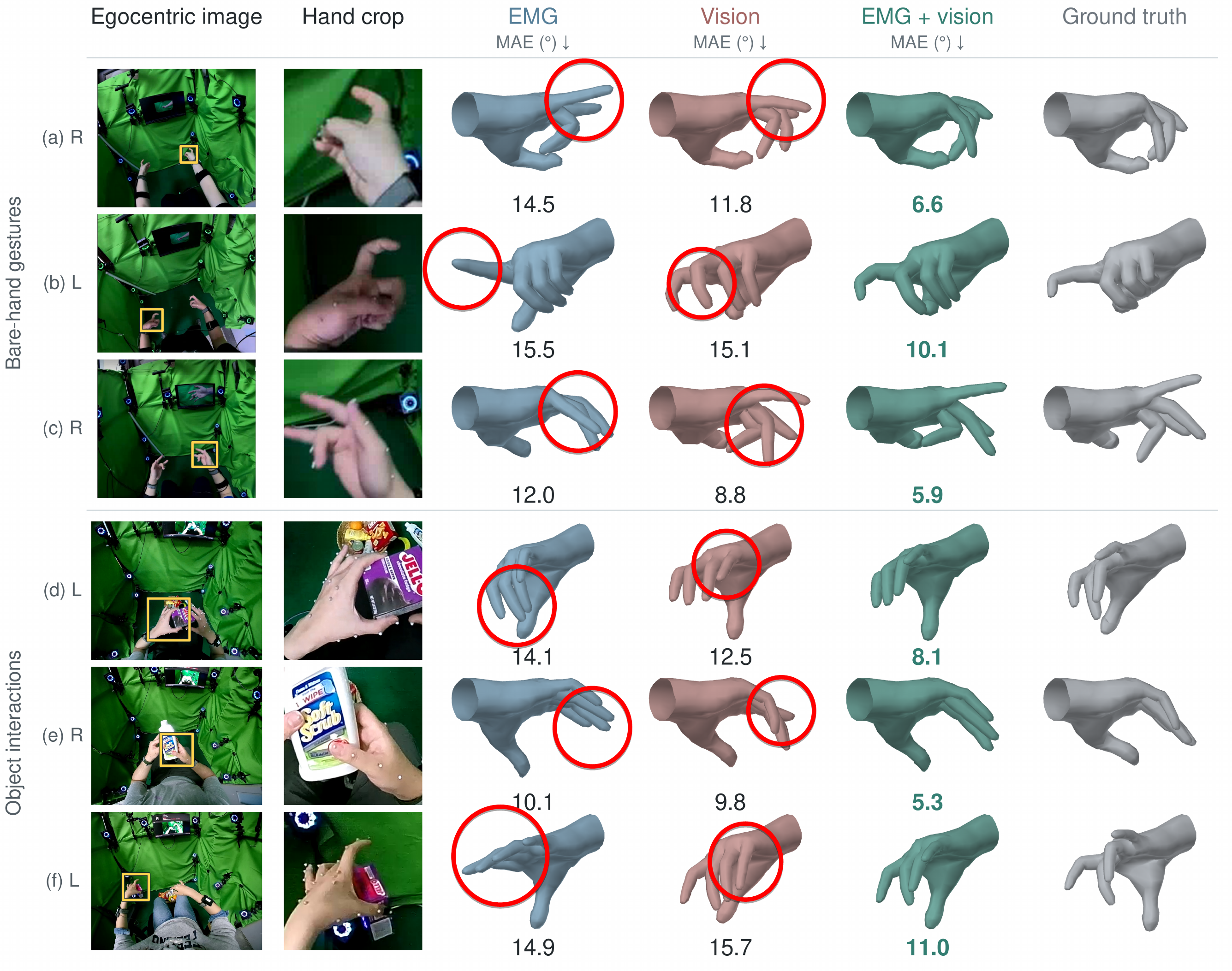}
\caption{Qualitative comparison on \dataset{} using \model{}-S, ViT-S/14,
and their fusion. (a--c) Bare-hand gestures; (d,f) pudding mix box;
(e) bleach cleanser bottle. L/R denotes the target hand. The boxed hand is
enlarged in the crop column for display; vision models use the full egocentric image.
Red circles highlight local pose errors.
Values indicate MAE over valid joint angles ($^\circ$, lower is better).}
\label{fig:qualitative_fusion}
\end{figure*}

\subsection{Fusion Design and Temporal Context}
\label{sec:temporal_context}

We compare direct and residual fusion of EMG and vision features
(Table~\ref{tab:temporal_results}). Residual fusion achieves a slightly
lower Overall MAE: $8.05^\circ$ versus $8.14^\circ$ for direct fusion.

Using 17 frames around the target frame reduces the ViT-S Overall MAE from
$8.55^\circ$ to $7.72^\circ$. Adding EMG further reduces it to
$7.47^\circ$, with lower errors in all four evaluation conditions.
Further EMG ablations and error breakdowns appear in
\supplementref{app:extended_results}.

\begin{table}[t]
\centering
\caption{Fusion design and temporal context ablations with ViT-S/14 (MAE, $^\circ$, $\downarrow$). Temporal models use 17 RGB frames over 4\,s. Shading highlights temporal fusion; bold marks the lowest error in each column.}
\label{tab:temporal_results}
\footnotesize
\setlength{\tabcolsep}{3pt}
\renewcommand{\arraystretch}{1.12}
\begin{tabularx}{\columnwidth}{l*{5}{>{\raggedleft\arraybackslash}X}}
\toprule
Model & Gesture & User & G+U & Object & Overall \\
\midrule
ViT-S & 7.68 & 8.74 & 9.09 & 9.02 & 8.55 \\
Direct fusion & 7.39 & 8.55 & 8.61 & 8.40 & 8.14 \\
Residual fusion & 7.36 & 8.33 & 8.47 & 8.34 & 8.05 \\
Temporal & 6.66 & 7.89 & 7.83 & 8.37 & 7.72 \\
\rowcolor{resultshade}
Temporal + EMG & $\mathbf{6.51}$ & $\mathbf{7.74}$ & $\mathbf{7.68}$ & $\mathbf{7.99}$ & \cellcolor{resultfocus}$\mathbf{7.47}$ \\
\bottomrule
\end{tabularx}
\end{table}

\subsection{Qualitative Analysis}
\label{sec:qualitative_analysis}

In Figure~\ref{fig:qualitative_fusion}(a), the EMG-only model
correctly predicts the articulation of the self-occluded middle
and ring fingers, but incorrectly predicts an extended little
finger. The vision-only model struggles to estimate the articulation
of the middle and ring fingers, whereas the fusion model correctly
recovers the articulation of all fingers. Similar patterns appear
in Figure~\ref{fig:qualitative_fusion}(b,c). In (c), the vision-only
model also struggles with motion blur caused by rapid hand movement,
while the fusion model estimates the pose more accurately.

In Figure~\ref{fig:qualitative_fusion}(d--f), objects occlude the
hands to varying degrees, limiting the visual evidence for pose
estimation. For example, in (e), the bottle occludes much of the
right hand, and the vision-only model incorrectly estimates the
articulation of all four occluded fingers. The fusion model
recovers their articulation more accurately. These examples
illustrate how EMG complements vision under both self-occlusion
and object occlusion.

\subsection{Fusion under External Occlusion}
\label{sec:occlusion_analysis}

\begin{figure}[t]
\centering
\includegraphics[width=\columnwidth]{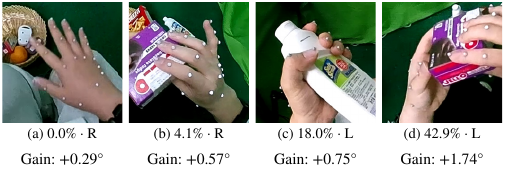}
\caption{Selected hand crops from the four occlusion groups in
Table~\ref{tab:occlusion_gain_bins}, ordered from lowest to highest.
Each panel reports the estimated external occlusion score, target hand (L/R),
and that sample's fusion gain (vision MAE minus fusion MAE, in degrees).}
\label{fig:occlusion_examples}
\end{figure}

We measure the projected hand area hidden by the opposite hand and
objects over 21 fixed wrist and finger regions on the MANO surface.
For region $r$, let $A_{0r}$ denote its isolated projected area,
$A_{hr}$ the area visible after self-occlusion, and $A_{sr}$ the area
remaining after occlusion by the opposite hand and objects. Averaging
equally over valid regions $\mathcal V$, we define
\begin{equation}
o_{\mathrm{ext}} = \frac{1}{|\mathcal V|}
\sum_{r\in\mathcal V}\frac{A_{hr}-A_{sr}}{A_{0r}}.
\label{eq:external_occlusion}
\end{equation}
The reported score is $100o_{\mathrm{ext}}\%$.
We compare ViT-S/14 and its \model{}-S fusion model on the Object test set.
Samples are grouped by quartiles of the estimated occlusion scores, with
zero-score samples kept together. We estimate 95\% confidence intervals
using 2{,}000 paired bootstrap resamples at the participant level.
Table~\ref{tab:occlusion_gain_bins} shows that fusion reduces MAE in all
four groups.
The absolute reduction increases from $0.59^\circ$ in the zero-score group
to $1.20^\circ$ in the highest-score group, a difference of $0.61^\circ$
(95\% CI: $[0.39,0.81]^\circ$).
Figure~\ref{fig:occlusion_examples} shows one selected example from each group.

\begin{table}[t]
\centering
\caption{MAE ($^\circ$) by estimated external occlusion for ViT-S/14 and
its \model{}-S fusion model, using only Object test samples.
$\Delta$ is vision MAE minus fusion MAE; brackets show 95\% confidence
intervals from participant-level bootstrap resampling. Differences are computed before rounding.}
\label{tab:occlusion_gain_bins}
\footnotesize
\setlength{\tabcolsep}{3pt}
\renewcommand{\arraystretch}{1.12}
\begin{tabular*}{\columnwidth}{@{\extracolsep{\fill}}lrrr@{}}
\toprule
Score (\%) & Vision & Fusion & $\Delta$ [95\% CI] \\
\midrule
$0$ & 7.65 & 7.06 & 0.59 [0.50, 0.70] \\
$(0,8.33]$ & 9.55 & 8.85 & 0.69 [0.59, 0.80] \\
$(8.33,29.81]$ & 11.03 & 10.21 & 0.81 [0.67, 0.96] \\
\rowcolor{resultshade}
$>29.81$ & 13.07 & 11.86 & 1.20 [0.96, 1.42] \\
\bottomrule
\end{tabular*}
\end{table}

\section{Discussion and Conclusion}

\dataset{} connects muscle activity and visual observations through
continuous hand-pose annotations. Shared targets and evaluation splits
support comparisons across participants, gestures, and objects, while
bilateral recordings capture coordinated hand movements and object
manipulation.

EMG remains useful alongside stronger visual backbones and temporal
image context. For ViT-S, adding EMG to the temporal model reduces MAE
from $7.72^\circ$ to $7.47^\circ$, suggesting complementary information
beyond neighboring RGB frames. On the Object test set, fusion gains
increase from $0.59^\circ$ to $1.20^\circ$ across the estimated occlusion
groups.

\paragraph{Limitations and future work.}
Generalization to unseen participants remains difficult for EMG-based
estimation.
This motivates cross-dataset pretraining and adaptation with short
recordings from a new user. The fusion model pools each EMG window into
one vector; future models could align both modalities throughout the
sequence and adapt their contributions to hand visibility. Object meshes
and pose annotations also support extending the benchmark to joint
hand--object motion and contact estimation.

\noindent\begin{minipage}{\columnwidth}
\section*{Ethics Statement}

Participants gave informed consent under an institutionally approved ethics protocol, received standard hourly compensation, and could withdraw and request data deletion; none did. Consent covered the collection purpose, recorded modalities, and intended research use.

Faces in egocentric and external videos are automatically blurred, and participant IDs are anonymized. MANO shape parameters and EMG signals may still enable biometric identification. The dataset will be released under CC-BY-NC 4.0 for noncommercial research, with guidance against re-identification. 

\end{minipage}\par

\section*{LLM Usage Statement}
OpenAI tools (ChatGPT/GPT and Codex) assisted with manuscript feedback, language, presentation, and formatting, including related work, experimental discussions, and claims. They also generated figure scripts, implementation code, and utilities, which the authors reviewed, edited, and verified. No LLMs were used in dataset creation, processing, or annotation. The authors take full responsibility for all content.

\FloatBarrier
{
\small
\bibliographystyle{ieeenat_fullname}
\bibliography{references}
}
\clearpage
\appendix
\setcounter{figure}{0}
\setcounter{table}{0}
\setcounter{equation}{0}
\renewcommand{\thefigure}{S\arabic{figure}}
\renewcommand{\thetable}{S\arabic{table}}
\renewcommand{\theequation}{S\arabic{equation}}
\renewcommand{\theHfigure}{appendix.\arabic{figure}}
\renewcommand{\theHtable}{appendix.\arabic{table}}
\renewcommand{\theHequation}{appendix.\arabic{equation}}
\section*{Appendix}
\suppressfloats[t]
\raggedbottom

\section{Additional Results}
\label{app:extended_results}

\subsection{Center-Frame EMG Evaluation}
\label{app:emg_center_test}

We evaluate the pretrained \model{}-S/M/L models at center frames sampled
every 400\,ms, retaining centers with valid EMG windows
(Table~\ref{tab:emg_center_test}).
Under this center-frame evaluation protocol, Overall MAE is lower than
under the full-window protocol in Table~\mainref{tab:egoemg_emg_results}
by $0.23^\circ$, $0.18^\circ$, and $0.18^\circ$ for S, M, and L, respectively.

\begin{table}[!htbp]
\centering
\caption{EMG-only center-frame MAE ($^\circ$, $\downarrow$) for \model{}-S/M/L. Overall pools valid angles across conditions and hands. Bold marks the lowest error per column.}
\label{tab:emg_center_test}
\footnotesize
\setlength{\tabcolsep}{3pt}
\begin{tabular*}{\columnwidth}{@{\extracolsep{\fill}}lrrrrr@{}}
\toprule
Model & Gesture & User & G+U & Object & Overall \\
\midrule
\model{}-S & \textbf{13.68} & 15.24 & 17.07 & \textbf{9.82} & \textbf{12.05} \\
\model{}-M & 13.69 & 15.72 & 17.55 & 10.06 & 12.28 \\
\model{}-L & 14.17 & \textbf{14.78} & \textbf{16.51} & 10.21 & 12.28 \\
\bottomrule
\end{tabular*}
\end{table}

\subsection{Causal EMGFormer}
\label{app:emg_model_causal}

We train causal variants of \model{}-S/M/L in which each prediction
depends only on EMG samples available up to that time. EMG filtering
uses causal filters, and each convolutional feature is assigned to the
last sample in its receptive field. Squeeze-and-excitation aggregates
features only up to and including the current time step, and Transformer
attention is masked to exclude future features. Between successive outputs, we hold the
latest available prediction.
Table~\ref{tab:emg_model_causal} compares causal and bidirectional models
on the same evaluation samples and valid joint angles.

Overall MAE changes by less than $0.17^\circ$ for each model size, while
User MAE increases for all three causal variants.

\begin{table}[t]
\centering
\caption{EMG-to-pose MAE ($^\circ$, $\downarrow$) on \dataset{} with bidirectional and causal models. Both are evaluated throughout the EMG window on the same valid angles. Causal variants use causally filtered EMG. Shading marks causal variants.}
\label{tab:emg_model_causal}
\footnotesize
\setlength{\tabcolsep}{3pt}
\renewcommand{\arraystretch}{1.12}
\resizebox{\columnwidth}{!}{%
\begin{tabular}{llrrrrr}
\toprule
Model & Context & Gesture & User & G+U & Object & Overall \\
\midrule
\model{}-S & Bidirectional & 13.40 & 15.37 & 16.87 & 9.92 & 12.28 \\
\rowcolor{resultshade}
\model{}-S & Causal & 13.18 & 15.67 & 16.38 & 9.84 & 12.23 \\
\addlinespace[2pt]
\model{}-M & Bidirectional & 13.40 & 15.69 & 17.19 & 10.15 & 12.46 \\
\rowcolor{resultshade}
\model{}-M & Causal & 13.06 & 15.96 & 17.10 & 9.90 & 12.30 \\
\addlinespace[2pt]
\model{}-L & Bidirectional & 13.92 & 14.83 & 16.14 & 10.29 & 12.46 \\
\rowcolor{resultshade}
\model{}-L & Causal & 13.52 & 16.18 & 17.16 & 9.98 & 12.52 \\
\bottomrule
\end{tabular}}%
\end{table}

\subsection{EMG Filtering and Augmentation}
\label{app:fusion_freeze_ablation}

Table~\ref{tab:fusion_freeze_ablation} reports Overall MAE for all combinations
of model size, EMG filtering, and training augmentation. Filtered EMG with
augmentation achieves the lowest Overall MAE for each model size.
For raw EMG, the effect of augmentation varies with model size.

\begin{table}[!htbp]
\centering
\caption{Effect of EMG filtering and training augmentation. Overall MAE ($^\circ$) for EMGFormer S/M/L; lower is better.}
\label{tab:fusion_freeze_ablation}
\footnotesize
\setlength{\tabcolsep}{3.5pt}
\renewcommand{\arraystretch}{1.12}
\begin{tabular}{@{}lcrrr@{}}
\toprule
Input & Aug. & \model{}-S & \model{}-M & \model{}-L \\
\midrule
Raw & -- & 14.03 & 14.10 & 14.14 \\
Raw & $\checkmark$ & 14.08 & 14.02 & 14.01 \\
\midrule
Filtered & -- & 12.50 & 12.61 & 12.56 \\
\rowcolor{gray!12}
Filtered & $\checkmark$ & $\mathbf{12.28}$ & $\mathbf{12.46}$ & $\mathbf{12.46}$ \\
\bottomrule
\end{tabular}
\end{table}

\subsection{ResNet Results}
\label{app:resnet_results}

Table~\ref{tab:resnet_results} reports ResNet-18/50/152 vision and fusion
results. Fusion reduces Overall MAE for all three ResNet backbones. ResNet-152 achieves the lowest vision-only error among the ResNets
($8.76^\circ$ Overall); ResNet-50 shows the largest reduction in Overall MAE after adding EMG
($0.55^\circ$).

\begin{table}[!htbp]
\centering
\begin{threeparttable}
\caption{ResNet vision and EMG+vision results (MAE, $^\circ$). $\Delta$ is the reduction in Overall MAE from fusion; bold marks the lowest error per column. Differences are computed before rounding.}
\label{tab:resnet_results}
\footnotesize
\setlength{\tabcolsep}{3.5pt}
\renewcommand{\arraystretch}{1.12}
\begin{tabular}{@{}llrrrrrr@{}}
\toprule
Backbone & Input & Gesture & User & G+U & Object & Overall & $\Delta$ \\
\midrule
ResNet-18 & Vision & 9.31 & 10.33 & 10.29 & 10.24 & 9.98 & -- \\
 & + EMG & 9.05 & 10.31 & 10.27 & 9.76 & 9.71 & +0.27 \\
\addlinespace[2pt]
ResNet-50 & Vision & 9.06 & 10.31 & 10.41 & 10.12 & 9.85 & -- \\
 & + EMG & 8.68 & 9.99 & 9.99 & 9.31 & 9.30 & +0.55 \\
\addlinespace[2pt]
ResNet-152 & Vision & 7.94 & 8.94 & 8.98 & 9.21 & 8.76 & -- \\
 & + EMG & $\mathbf{7.75}$ & $\mathbf{8.84}$ & $\mathbf{8.80}$ & $\mathbf{8.70}$ & $\mathbf{8.45}$ & +0.31 \\
\bottomrule
\end{tabular}
\end{threeparttable}
\end{table}

\subsection{Errors by Gesture and Object}
\label{app:per_gesture}

Fusion gains vary across gestures (Table~\ref{tab:per_gesture_full}).
For example, ASL 4 and index bending show reductions of $1.68^\circ$
and $1.47^\circ$, respectively, whereas left-hand finger pulling shows
no improvement over vision alone.

\begin{table*}[t]
\centering
\caption{MAE by gesture or object ($^\circ$) for ViT-S/14 vision, EMGFormer-S and their fusion. $\Delta$ is Vision minus Fusion. EMG is evaluated throughout the EMG window; vision and fusion are evaluated at the center frame. $^\dagger$: interaction with the pudding mix box. Differences are computed before rounding. Free bimanual motion is omitted because results are unavailable for some methods.}
\label{tab:per_gesture_full}
\footnotesize
\setlength{\tabcolsep}{2pt}
\renewcommand{\arraystretch}{1.0}
\begin{minipage}[t]{0.48\textwidth}\centering
\begin{tabularx}{\linewidth}{@{}>{\raggedright\arraybackslash}Xrrrr@{}}
\toprule
Gesture / object & Vision & EMG & Fusion & $\Delta$ \\
\midrule
Pudding mix box$^\dagger$ & 9.02 & 9.92 & 8.34 & +0.68 \\
ASL 1 & 9.20 & 19.75 & 9.01 & +0.18 \\
ASL 2 & 9.11 & 16.40 & 8.35 & +0.76 \\
ASL 3 & 9.91 & 18.98 & 9.28 & +0.62 \\
ASL 4 & 9.07 & 15.92 & 7.39 & +1.68 \\
ASL 5 & 7.16 & 11.94 & 5.89 & +1.27 \\
ASL 6 & 10.27 & 16.78 & 10.04 & +0.23 \\
ASL 7 & 8.25 & 14.41 & 7.65 & +0.59 \\
ASL 8 & 8.62 & 13.78 & 7.77 & +0.85 \\
ASL 9 & 8.10 & 18.05 & 7.35 & +0.75 \\
Three-finger claw & 9.70 & 17.83 & 9.60 & +0.10 \\
Five-finger claw & 8.47 & 15.02 & 7.98 & +0.49 \\
Free hand motion & 7.71 & 12.69 & 7.36 & +0.36 \\
ILY & 8.77 & 18.07 & 8.44 & +0.33 \\
Index bend & 9.96 & 17.16 & 8.49 & +1.47 \\
Two-finger claw & 10.52 & 19.90 & 10.12 & +0.40 \\
Thumb circles & 6.83 & 11.97 & 6.49 & +0.33 \\
Thumb sliding & 6.97 & 18.99 & 6.23 & +0.74 \\
Middle bend & 9.19 & 14.37 & 8.73 & +0.46 \\
Number 9 & 9.27 & 15.52 & 8.72 & +0.55 \\
Palm rotation & 8.27 & 11.05 & 7.21 & +1.06 \\
Middle-finger pinch & 9.04 & 12.30 & 8.41 & +0.62 \\
Little-finger bend & 12.88 & 15.78 & 12.11 & +0.77 \\
Rest & 9.05 & 19.04 & 8.91 & +0.14 \\
Ring-finger pinch & 9.64 & 16.65 & 9.15 & +0.49 \\
Ring-finger bend & 7.67 & 14.82 & 7.33 & +0.34 \\
Rock sign & 8.74 & 15.45 & 8.40 & +0.33 \\
Thumbs up & 9.23 & 26.25 & 9.21 & +0.01 \\
Open palm & 6.79 & 14.16 & 6.47 & +0.31 \\
Free finger motion & 9.37 & 13.98 & 9.23 & +0.14 \\
\bottomrule\end{tabularx}
\end{minipage}\hfill
\begin{minipage}[t]{0.48\textwidth}\centering
\begin{tabularx}{\linewidth}{@{}>{\raggedright\arraybackslash}Xrrrr@{}}
\toprule
Gesture / object & Vision & EMG & Fusion & $\Delta$ \\
\midrule
Grasping in air & 10.03 & 18.62 & 9.63 & +0.40 \\
Clap & 8.89 & 16.31 & 8.41 & +0.47 \\
Crossed fingers & 6.63 & 11.65 & 6.15 & +0.48 \\
Crossed stretch & 6.24 & 11.09 & 6.14 & +0.10 \\
Left-hand finger pull & 9.41 & 15.46 & 9.53 & -0.11 \\
Right-hand finger pull & 9.58 & 13.34 & 9.45 & +0.13 \\
Fingertip touch & 7.38 & 12.45 & 7.19 & +0.20 \\
Fist bump & 8.13 & 19.18 & 7.88 & +0.25 \\
Gaming & 9.51 & 17.62 & 9.33 & +0.18 \\
Hand clasp & 10.32 & 15.82 & 9.77 & +0.56 \\
Palm rubbing & 6.96 & 11.33 & 6.59 & +0.37 \\
Index tapping & 7.40 & 11.38 & 7.37 & +0.03 \\
Palm rolling & 5.30 & 11.19 & 5.12 & +0.18 \\
Stacked palms & 4.30 & 8.88 & 4.11 & +0.19 \\
Hooked little fingers & 9.47 & 14.87 & 8.97 & +0.49 \\
Prayer & 7.25 & 12.10 & 7.14 & +0.11 \\
Squeeze & 7.51 & 14.01 & 7.11 & +0.40 \\
Symmetric opening & 7.09 & 12.11 & 6.98 & +0.12 \\
Symmetric swinging & 8.19 & 10.70 & 8.03 & +0.16 \\
Thumb wrestling & 6.79 & 11.19 & 6.69 & +0.10 \\
Typing & 7.73 & 11.69 & 7.73 & 0.00 \\
Finger touch & 7.50 & 13.44 & 7.51 & -0.01 \\
Middle-finger opposition & 5.33 & 10.91 & 5.12 & +0.21 \\
``Bei'' sign & 12.63 & 19.36 & 12.37 & +0.26 \\
Thumb sign & 13.96 & 22.01 & 13.22 & +0.73 \\
Two-hand claw & 8.90 & 16.93 & 8.59 & +0.31 \\
Two-hand OK & 8.67 & 14.47 & 8.49 & +0.18 \\
Picture frame & 10.69 & 18.37 & 10.03 & +0.66 \\
Pinch and twist & 8.60 & 16.22 & 8.20 & +0.40 \\
Thumb opposition & 9.57 & 16.35 & 9.31 & +0.27 \\
\bottomrule\end{tabularx}
\end{minipage}
\end{table*}

\subsection{Errors by Hand and Joint}
\label{app:emg_per_hand}
\label{app:per_joint}

Right-hand Overall MAE is lower than left-hand MAE for all three model
sizes (Table~\ref{tab:emg_per_hand}). Across joint angles, little-finger
MCP flexion has the highest error for each model, while ring-finger MCP
abduction has the lowest (Table~\ref{tab:per_joint_egoemg}).

\begin{table}[!htbp]
\centering
\caption{Per-hand EMG-to-pose MAE ($^\circ$), averaged over valid joint angles throughout the EMG windows.}
\label{tab:emg_per_hand}
\footnotesize
\setlength{\tabcolsep}{3pt}
\renewcommand{\arraystretch}{1.12}
\begin{tabular*}{\columnwidth}{@{\extracolsep{\fill}}lrrrrr@{}}
\toprule
Method & Gesture & User & G+U & Object & Overall \\
\midrule
\model{}-S / Left & 13.54 & 15.51 & 16.97 & 10.13 & 12.45 \\
\model{}-S / Right & 13.25 & 15.22 & 16.77 & 9.71 & 12.10 \\
\model{}-M / Left & 13.68 & 15.62 & 16.83 & 10.49 & 12.68 \\
\model{}-M / Right & 13.10 & 15.75 & 17.55 & 9.81 & 12.24 \\
\model{}-L / Left & 14.11 & 14.75 & 15.89 & 10.54 & 12.62 \\
\model{}-L / Right & 13.73 & 14.92 & 16.39 & 10.03 & 12.31 \\
\bottomrule
\end{tabular*}
\end{table}

\begin{table}[t]
\centering
\caption{Per-angle EMGFormer MAE ($^\circ$), pooled over all four evaluation conditions and both hands. Joint abbreviations follow Table~\ref{tab:joint_angle_semantics}.}
\label{tab:per_joint_egoemg}
\scriptsize
\setlength{\tabcolsep}{3pt}
\renewcommand{\arraystretch}{1.0}
\begin{tabular*}{\columnwidth}{@{\extracolsep{\fill}}lrrr@{}}
\toprule
Joint & \model{}-S & \model{}-M & \model{}-L \\
\midrule
Thumb CMC FE & 8.96 & 9.18 & 9.24 \\
Thumb CMC AA & 12.13 & 12.60 & 12.56 \\
Thumb MCP FE & 11.12 & 11.54 & 11.49 \\
Thumb IP FE & 11.25 & 11.65 & 11.67 \\
Index MCP AA & 6.93 & 7.00 & 7.09 \\
Index MCP FE & 14.16 & 14.22 & 14.13 \\
Index PIP FE & 17.67 & 17.67 & 17.74 \\
Index DIP FE & 8.12 & 8.07 & 8.12 \\
Middle MCP AA & 5.41 & 5.54 & 5.50 \\
Middle MCP FE & 15.63 & 16.35 & 16.01 \\
Middle PIP FE & 18.53 & 18.68 & 18.53 \\
Middle DIP FE & 10.75 & 10.84 & 10.81 \\
Ring MCP AA & 5.25 & 5.28 & 5.32 \\
Ring MCP FE & 16.39 & 16.93 & 16.86 \\
Ring PIP FE & 20.19 & 20.40 & 20.37 \\
Ring DIP FE & 7.83 & 7.86 & 7.89 \\
Little MCP AA & 8.80 & 8.79 & 8.92 \\
Little MCP FE & 21.01 & 21.18 & 21.30 \\
Little PIP FE & 18.95 & 19.26 & 19.43 \\
Little DIP FE & 9.15 & 9.15 & 9.37 \\
Wrist FE & 11.67 & 11.73 & 11.69 \\
Wrist deviation & 9.85 & 9.93 & 9.80 \\
\bottomrule
\end{tabular*}
\end{table}

\section{Dataset Documentation}
\label{app:dataset_details}

\subsection{Dataset Card}
\label{app:dataset_card}

\paragraph{License.}
The licenses are CC-BY-NC 4.0 for research use of the dataset and MIT for the baseline code. Third-party assets, including the MANO model~\citep{mano} and the training data for \marker{}, remain subject to their original licenses.

\paragraph{Data organization.}
The dataset index lists 51 bare-hand and 30 HOI sessions. Each session contains
calibration metadata and separate hand streams with timestamps, raw and
filtered EMG, IMU, MANO pose and shape, UmeTrack finger angles, wrist
angles, validity flags, and split assignments. Camera streams contain
MP4 videos and timestamps, linked to hand measurements by stored frame
correspondences and time differences.
Depth streams are stored separately from RGB, with their own timestamps
and validity flags. External depth is provided by ZED and
RealSense cameras; the two ZED models are grouped together, and the two
RealSense cameras are treated as a pair in the availability statistics.

\paragraph{Egocentric cameras.}
We use a GoPro 13 for gesture recordings and a Dexwise head-mounted
camera for object-interaction recordings. The GoPro provides RGB video;
the Dexwise additionally records a monochrome stereo pair for depth
estimation.
Stored GoPro 13 and Dexwise videos have resolutions of $1280\times720$
and $1600\times1296$, respectively, both at 60\,fps.

\paragraph{Temporal alignment.}
All modalities are expressed on a common host-clock timeline. EMG and IMU
timestamps are reconstructed from serial-data arrival times and sample
order. Camera frames use SDK timestamps where available; Dexwise videos
use recording timestamps aligned to the host clock. Motion-capture
timestamps are mapped from the capture-system clock to the host timeline.
For RealSense recordings with paired RGB and depth frames, depth timestamps
are taken from the corresponding RGB frames.

Measurements are matched to the EMG timeline using zero-order holding for
RGB and depth frames and linear interpolation for motion-capture-derived
pose measurements. For valid hand-sample/egocentric-RGB matches, the median
and 95th-percentile absolute differences between sample and frame
timestamps are 7.10 and 16.05\,ms, respectively.

\paragraph{Held-out gestures.}
\label{app:split_details}
The held-out gestures are Thumb circles, Number 9, Middle-finger pinch,
Free finger motion, Fist bump, Stacked palms, Squeeze, and
Middle-finger opposition.
Participant counts and the validation/test protocol are given in
Section~\mainref{sec:data_partitioning}.

\paragraph{Data availability.}
Table~\ref{tab:modality_coverage} reports coverage among 356.61 million
single-hand samples assigned to training, validation, or testing.
Each sample represents one hand at one timestamp.

EMG and IMU samples require finite values in every channel and valid
entries in the stored validity masks, where present. Pose-label coverage
requires all 22 finger and wrist angles to be finite and marked valid.

RGB coverage requires an available video, a valid corresponding frame
and timestamp, and an absolute sample-to-frame time difference of at most
50\,ms. Joint EMG + RGB + pose-label coverage requires all three to be
valid at the same hand-sample timestamp.

Depth coverage requires a stored frame with a valid timestamp and valid
entries in the supplied masks. The most recent depth frame at or before
the hand-sample timestamp must be at most 100\,ms old, matching the
released 10\,Hz depth rate. RealSense coverage requires both cameras to
satisfy these criteria.

\begin{table*}[t]
\centering
\caption{Availability of sensor measurements and joint-angle labels (\%) among samples assigned to training or evaluation. Each sample describes one hand at one time point. Percentages use the sample count shown in each column.}
\label{tab:modality_coverage}
\footnotesize
\setlength{\tabcolsep}{6pt}
\begin{tabular*}{\textwidth}{@{\extracolsep{\fill}}lrrrrrr@{}}
\toprule
Measurements / labels & All & Bare-hand & HOI & Train & Validation & Test \\
\midrule
EMG (8 channels) & 100.00 & 100.00 & 100.00 & 100.00 & 100.00 & 100.00 \\
IMU (6 axes) & 99.54 & 99.02 & 100.00 & 99.52 & 99.66 & 99.60 \\
Egocentric RGB & 97.73 & 97.49 & 97.94 & 97.54 & 98.60 & 98.32 \\
External depth (ZED) & 98.31 & 97.08 & 99.40 & 98.39 & 98.13 & 97.90 \\
External depth (RealSense pair) & 37.92 & 0.00 & 71.20 & 37.29 & 42.82 & 37.75 \\
Pose labels (22 angles) & 91.40 & 94.92 & 88.30 & 90.69 & 93.02 & 95.20 \\
\midrule
EMG + RGB + pose labels & 89.71 & 92.63 & 87.14 & 88.86 & 91.99 & 93.85 \\
\midrule
Single-hand samples ($10^6$) & 356.61 & 166.66 & 189.95 & 282.21 & 37.68 & 36.73 \\
\bottomrule
\end{tabular*}
\end{table*}

\subsection{Hand--Object Interaction Coverage}
\label{app:hoi_coverage}

\begin{figure}[!htbp]
\centering
\includegraphics[width=\columnwidth,trim=0 0 0 0,clip]{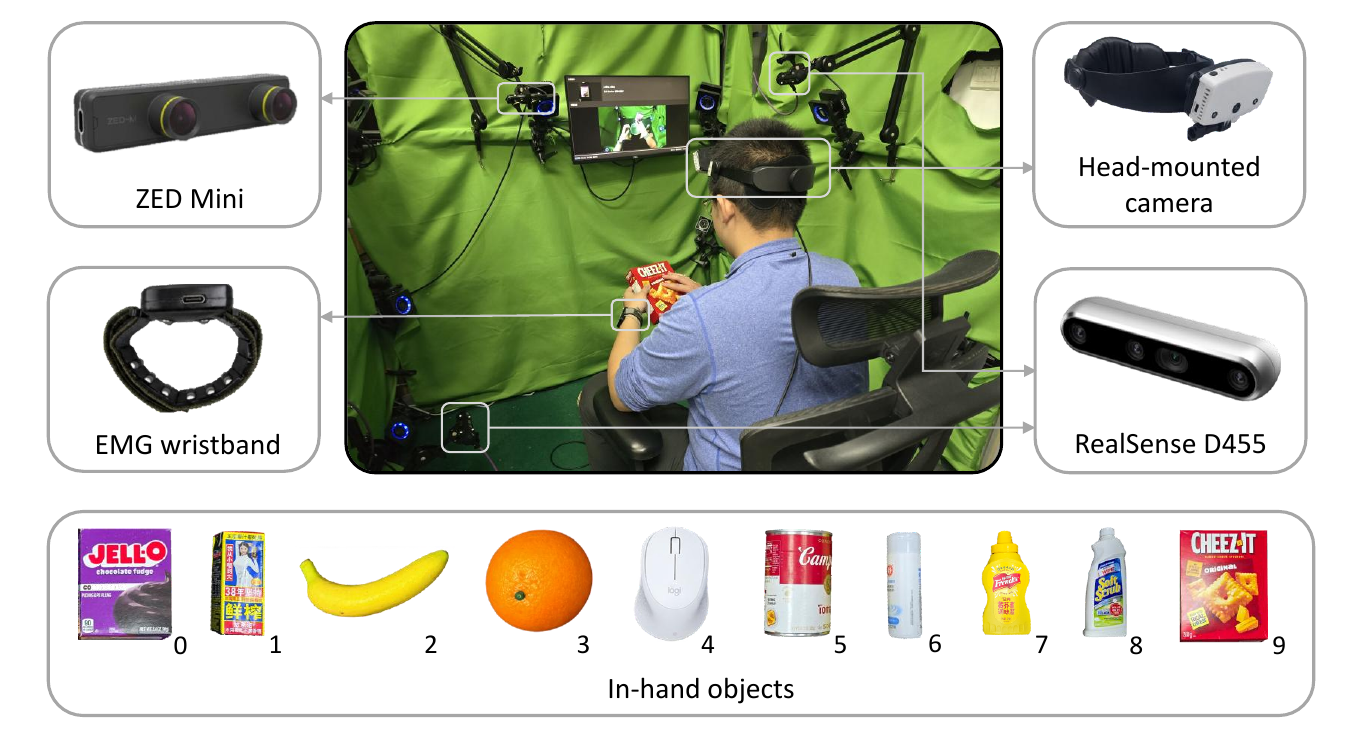}
\caption{HOI capture setup and object set. One ZED Mini and two RealSense D455 cameras recorded HOI sessions, providing extra RGB-D views for object pose estimation and refinement.}
\label{fig:hoi_capture_setup}
\end{figure}

The HOI subset contains 30 sessions totaling 14.13\,hours from 16 participants. Nine participants were assigned tasks involving all 10 objects; the others were assigned a subset. Task labels identify the object participants were asked to use. We provide object poses and scanned 3D meshes for all 10 objects (Table~\ref{tab:hoi_objects}; Figure~\ref{fig:hoi_capture_setup}), along with segmentation masks generated using SAM~3~\citep{sam3} for objects in egocentric images.

\begin{table}[t]
\centering
\caption{Objects in the annotated hand--object interaction subset.}
\label{tab:hoi_objects}
\begin{tabular}{@{}rl@{}}
\toprule
ID & Object \\
\midrule
0 & Pudding mix box \\
1 & Coconut milk carton \\
2 & Banana \\
3 & Orange \\
4 & Computer mouse \\
5 & Tomato soup can \\
6 & Facial cleanser bottle \\
7 & Mustard bottle \\
8 & Bleach cleanser bottle \\
9 & Cracker box \\
\bottomrule
\end{tabular}
\end{table}

\subsection{Gesture Vocabulary}
\label{app:gesture_vocabulary}

Table~\ref{tab:gesture_vocabulary} lists the 60 classes of the gesture subset. Object categories are listed in \supplementref{app:hoi_coverage}.

\begin{table*}[t]
\centering
\scriptsize
\setlength{\tabcolsep}{3pt}
\renewcommand{\arraystretch}{0.98}
\caption{Gesture vocabulary and descriptions for \dataset{} (60 classes).}
\label{tab:gesture_vocabulary}
\begin{tabular}{@{}p{0.13\textwidth}p{0.31\textwidth}p{0.13\textwidth}p{0.31\textwidth}@{}}
\toprule
\textbf{Gesture} & \textbf{Description} & \textbf{Gesture} & \textbf{Description} \\
\midrule
\multicolumn{2}{l}{\textit{Single-hand (30 gestures)}} &
\multicolumn{2}{l}{\textit{Symmetric bimanual (18 gestures)}} \\
\addlinespace[2pt]
ASL 1--ASL 9 & American Sign Language digits 1 through 9 &
Clap & Symmetric clapping \\
Three-finger claw & Three-finger claw pose (index + middle + ring flexed) &
Crossed fingers & Interlaced fingers of both hands \\
Five-finger claw & Five-finger claw pose (all fingers flexed) &
Crossed stretch & Fingers of opposite hands crossed and extended \\
Free hand motion & Unconstrained hand movement &
Fingertip touch & Matching fingertips of both hands touching \\
ILY & ``I Love You'' hand sign (thumb, index, little finger extended) &
Fist bump & Fists touching in a fist bump \\
Index bend & Index finger bending and extending &
Gaming & Bare-handed imitation of game-controller use \\
Two-finger claw & Index and middle fingers clawed, others extended &
Hand clasp & Hands clasped together \\
Thumb circles & Thumb moving in circles over the palm &
Palm rubbing & Rubbing palms together \\
Thumb sliding & Thumb sliding across the palm &
Index tapping & Both index fingers tapping in the air \\
Middle bend & Middle finger bending and extending &
Finger touch & Thumb and middle fingertips touching, then separating \\
Number 9 & Hand forming number 9 (four fingers curled, index up) &
Stacked palms & One palm resting on the other \\
Palm rotation & Rotation between palm-up and palm-down orientations &
Prayer & Palms pressed together in prayer position \\
Middle-finger pinch & Pinching with the thumb and middle finger &
Middle-finger opposition & Middle fingertips of the two hands touching \\
Little-finger bend & Little finger bending and extending &
Symmetric opening & Both hands opening symmetrically \\
Rest & Relaxed hand in neutral resting posture &
Symmetric swinging & Both hands swinging symmetrically side to side \\
Ring-finger pinch & Ring finger touching thumb tip &
Thumb wrestling & Thumbs wrestling each other \\
Ring-finger bend & Ring finger bending and extending &
Typing & Both hands typing on a virtual keyboard \\
Rock sign & ``Rock on'' hand sign (index and little finger extended) &
Free bimanual motion & Unconstrained bimanual hand movement \\
Thumbs up & Thumb extended upward &
\multicolumn{2}{l}{\textit{Asymmetric bimanual (12 gestures)}} \\
Open palm & Fingers spread apart with palm open, no contact &
Left-hand finger pull & Left hand pulls right-hand fingers \\
Free finger motion & Free-form finger motion without contact &
Right-hand finger pull & Right hand pulls left-hand fingers \\
Grasping in air & Simulated grasping motion without contact &
Palm rolling & Palms facing, rolling around each other \\
& & Hooked little fingers & Little fingers hooked together \\
& & Squeeze & Both hands squeezing an imaginary object \\
& & ``Bei'' sign & Hands crossed to form the Chinese character ``Bei'' \\
& & Thumb sign & Both thumbs raised with the remaining fingers curled \\
& & Two-hand claw & Both hands clawing with asymmetric finger configurations \\
& & Two-hand OK & Both hands forming OK signs \\
& & Picture frame & Hands forming a rectangular picture frame \\
& & Pinch and twist & Two hands pinching and wringing \\
& & Thumb opposition & Thumbs in opposition at different orientations \\
\bottomrule
\end{tabular}
\end{table*}

\subsection{EMG Signal Processing}
\label{app:emg_processing}

All EMG signals are recorded at 2\,kHz. Zero-phase filtering is applied to each session using a
20\,Hz high-pass, 450\,Hz low-pass, and notches at 50\,Hz harmonics
through 450\,Hz. Causal \model{} variants use causally filtered inputs
(\supplementref{app:emg_model_causal}).

\section{Pose Reconstruction Quality}
\label{app:label_validation}

\subsection{Finger Angles from UmeTrack Surface Fitting}
\label{app:joint_angle_conversion}

\dataset{} provides 20 scalar finger joint angles in the format used by
EMG2Pose and UmeTrack~\citep{emg2pose,umetrack}. We estimate these angles
by fitting fixed surface-marker attachments on the UmeTrack mesh directly
to the observed 3D markers.

\paragraph{Surface correspondences.}
Each of the 21 markers is attached to a mesh triangle with fixed barycentric
weights. The triangle indices and weights are shared across frames for each
hand. Skinning the mesh with finger angles $\mathbf{a}$ gives the surface
position of marker $i$. We reflect these positions to match the hand's
handedness, yielding $\mathbf{q}_i(\mathbf{a})$, then transform them into
world coordinates using positive scale $s$, rotation $R$, and translation
$\mathbf{t}$.

\paragraph{Optimization.}
We exclude the second marker (MANO vertex index 88;
\supplementref{app:marker_selection}) from fitting and residual evaluation
because imprecise physical placement can introduce calibration error. For each frame, we fit the remaining valid observed
markers $\mathbf{y}_i$ by minimizing
\begin{equation}
\mathcal{L}=\frac{1}{3|\mathcal{V}|}\sum_{i\in\mathcal{V}}\sum_{c=1}^{3}
\rho_{3\mathrm{mm}}\!\left([sR\mathbf{q}_i(\mathbf{a})+\mathbf{t}-\mathbf{y}_i]_c\right)
+\frac{10^{-4}}{20}\|\mathbf{a}\|_2^2,
\end{equation}
where $\mathcal{V}$ denotes the valid markers, $c$ indexes the three spatial
coordinates, and $\rho$ is the SmoothL1 loss with a transition threshold of
3\,mm. Residual coordinates are expressed in millimetres. We optimize the
finger angles, global wrist rotation and translation, and scale subject to
the UmeTrack joint-angle bounds. We compute the skinned marker positions
and their analytic Jacobians and minimize the objective with a trust-region
reflective solver. For each frame, we run the solver from both a temporal
warm start and a rigidly aligned rest pose, then retain the solution with
the lower objective value. Frames with fewer than six valid markers or
degenerate marker geometry are marked invalid.

\paragraph{Fitting residuals.}
Across 27,986,073 valid fitted hand frames from 81 recording sessions,
the mean per-frame root-mean-square Euclidean distance between fitted
and observed markers is 2.8\,mm. We summarize finite residuals at the
native motion-capture timestamps, before temporal interpolation or alignment
to the EMG samples.

\paragraph{Output.}
The prediction target contains 20 fitted finger angles (indices 0--19) and
two armband-referenced wrist angles computed from motion-capture markers
(indices 20--21; \supplementref{app:wrist_angles}).
Table~\ref{tab:joint_angle_semantics} defines each output dimension.

\begin{table}[t]
\centering
\caption{Definitions of the 22 pose targets: 20 finger joint angles and two armband-referenced wrist angles (indices 20--21; \supplementref{app:wrist_angles}). Each dimension is reported in degrees. FE = flexion/extension; AA = abduction/adduction; CMC = carpometacarpal; MCP = metacarpophalangeal; IP = interphalangeal; PIP/DIP = proximal/distal interphalangeal.}
\label{tab:joint_angle_semantics}
\footnotesize
\setlength{\tabcolsep}{2.5pt}
\renewcommand{\arraystretch}{1.02}
\begin{tabular}{clll}
\toprule
Index & Finger & Joint & Motion \\
\midrule
0 & Thumb & CMC & FE \\
1 & Thumb & CMC & AA \\
2 & Thumb & MCP & FE \\
3 & Thumb & IP & FE \\
4 & Index & MCP & AA \\
5 & Index & MCP & FE \\
6 & Index & PIP & FE \\
7 & Index & DIP & FE \\
8 & Middle & MCP & AA \\
9 & Middle & MCP & FE \\
10 & Middle & PIP & FE \\
11 & Middle & DIP & FE \\
12 & Ring & MCP & AA \\
13 & Ring & MCP & FE \\
14 & Ring & PIP & FE \\
15 & Ring & DIP & FE \\
16 & Little & MCP & AA \\
17 & Little & MCP & FE \\
18 & Little & PIP & FE \\
19 & Little & DIP & FE \\
20 & -- & Wrist & Flexion/Extension \\
21 & -- & Wrist & Radial/Ulnar deviation \\
\bottomrule
\end{tabular}
\end{table}

\subsection{Wrist Angles from Armband Markers}
\label{app:wrist_angles}

We derive two armband-referenced wrist angles for flexion/extension
and radial/ulnar deviation from motion-capture markers. Four armband markers
define the reference frame.

\paragraph{Reference directions.}
For each hand, let $\mathbf m_{it}$ denote armband markers in the
session-specific order and $R_t$ map the armband frame to world coordinates.
With $\operatorname{unit}(\mathbf v)=\mathbf v/\|\mathbf v\|_2$, define
\begin{equation}
\begin{aligned}
\hat{\mathbf F}&=\operatorname{unit}\!\left(
\operatorname{median}_t\operatorname{unit}\!\left[
R_t^\top(\mathbf m_{1t}-\mathbf m_{2t})\right]\right),\\
\hat{\mathbf L}&=\operatorname{unit}\!\left(
\operatorname{median}_t\operatorname{unit}\!\left[
R_t^\top(\mathbf m_{1t}-\mathbf m_{0t})\right]\right).
\end{aligned}
\label{eq:wrist_directions}
\end{equation}
The componentwise median is computed over valid reference time steps.
The plane normal is
\begin{equation}
\hat{\mathbf N}=s\,\operatorname{unit}
(\hat{\mathbf F}\times\hat{\mathbf L}),
\label{eq:wrist_normal}
\end{equation}
where $s=+1$ for the right hand and $s=-1$ for the left.

\paragraph{Angle convention.}
Let $\hat{\mathbf H}=\operatorname{unit}[R^\top(\mathbf k_9-\mathbf k_0)]$
be the direction from the wrist to the middle-finger MCP in the armband
frame (indices 0 and 9 in the 21-point hand ordering; time subscripts
omitted). The projection-based labels are
\begin{equation}
\begin{aligned}
\theta_{\mathrm{fe}}&=\arcsin(\hat{\mathbf H}\cdot\hat{\mathbf N}),\\
\theta_{\mathrm{ru}}&=\operatorname{atan2}
(\hat{\mathbf H}\cdot\hat{\mathbf L},\,
 \hat{\mathbf H}\cdot\hat{\mathbf F}).
\end{aligned}
\label{eq:wrist_angles}
\end{equation}
The armband plane defines zero flexion/extension. Zero deviation satisfies
$\hat{\mathbf H}\cdot\hat{\mathbf L}=0$ and
$\hat{\mathbf H}\cdot\hat{\mathbf F}>0$.
The formulation accommodates non-orthogonal reference directions $\hat{\mathbf F}$ and $\hat{\mathbf L}$.
The marker ordering sets the extension and radial-deviation sign convention.
We clip the arcsine input to $[-1,1]$, store angles in degrees,
and convert them to radians for training.
Table~\ref{tab:joint_angle_semantics} gives the output order.

\subsection{MANO Reconstruction}
\label{app:markers2mano}

\marker{} estimates MANO parameters from optical motion-capture markers.

\subsubsection{Marker Selection}
\label{app:marker_selection}

We select 21 vertices from the MANO mesh surface as virtual markers, distributed across the wrist, fingers, and palm: indices 191, 88, 253, 708, 729 (wrist/palm), 144, 87, 295, 319 (thumb), 220, 365, 407, 445 (index/middle/ring), and 183, 477, 518, 556, 83, 589, 635, 673 (little finger/palm). These indices are consistent across all MANO shape instances $\boldsymbol{\beta}$, enabling direct correspondence between captured 3D marker positions and the canonical MANO mesh.

\subsubsection{Transformer Architecture}

\begin{figure*}[t]
\centering
\includegraphics[width=0.9\textwidth,trim=0 0 0 22,clip]{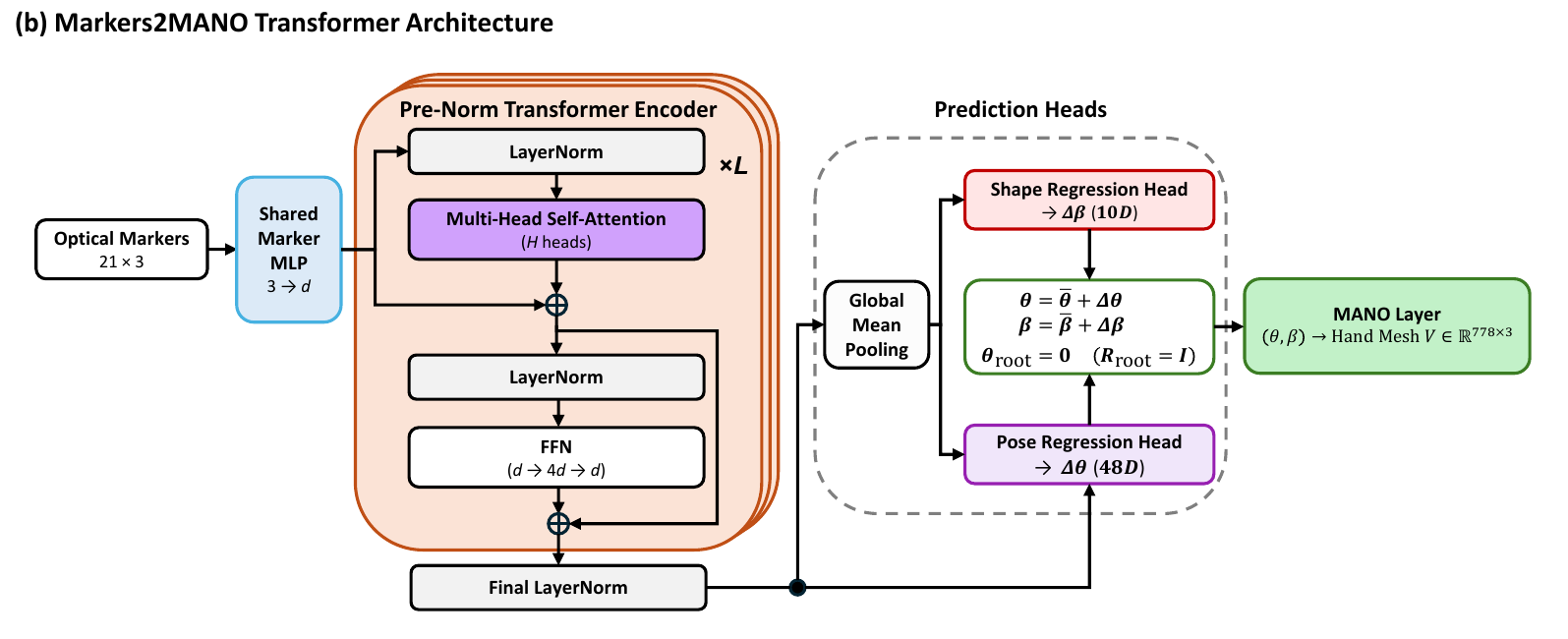}
\caption{\marker{} architecture for estimating MANO pose and shape from motion-capture markers.}
\label{fig:markers2mano_architecture}
\end{figure*}

We transform markers into a wrist-centered coordinate system with global
rotation removed and mirror left hands to match the right-hand convention. After reconstruction, we reverse
these transforms to recover the mesh in world coordinates and project it
into the image using camera calibration.

The Transformer in Figure~\ref{fig:markers2mano_architecture} maps
21 marker positions $\mathbf{M}\in\mathbb{R}^{21\times3}$ to MANO pose
$\boldsymbol{\theta}\in\mathbb{R}^{48}$ and shape
$\boldsymbol{\beta}\in\mathbb{R}^{10}$. It has $L=16$ pre-norm layers,
embedding dimension $d=1600$, 8 attention heads, feed-forward width
$4d=6400$, dropout 0.1, and 499.87M trainable parameters.

\begin{itemize}[leftmargin=1.5em,nosep]
    \item \textbf{Input embedding}: The model accepts all 21 transformed markers
    and removes five redundant
    root/MCP markers (indices 1, 5, 9, 13, and 17 in marker order), retaining
    16 tokens. A shared linear layer projects each 3D position to a
    $d$-dimensional token ($\mathbb{R}^3 \rightarrow \mathbb{R}^d$).
    \item \textbf{Output heads}: After a final LayerNorm, global mean pooling
    feeds an MLP shape head that predicts
    $\Delta\boldsymbol{\beta}\in\mathbb{R}^{10}$. For pose, each retained
    token is concatenated with the global feature and passed through a shared
    MLP that predicts a 6D rotation for one of the 16 MANO joints. The predicted
    pose and shape residuals are added to means computed from valid training
    labels. The heads are initialized to predict zero residuals, and the root rotation is
    fixed to identity.
    \item \textbf{MANO forward pass}: The predicted $\boldsymbol{\theta}$ and $\boldsymbol{\beta}$ are passed through the MANO layer to obtain the final mesh vertices $\mathbf{V} \in \mathbb{R}^{778 \times 3}$.
\end{itemize}

\subsubsection{Training Details}

\begin{figure*}[t]
\centering
\begingroup
\setlength{\fboxsep}{0pt}
\begin{overpic}[width=0.9\textwidth]{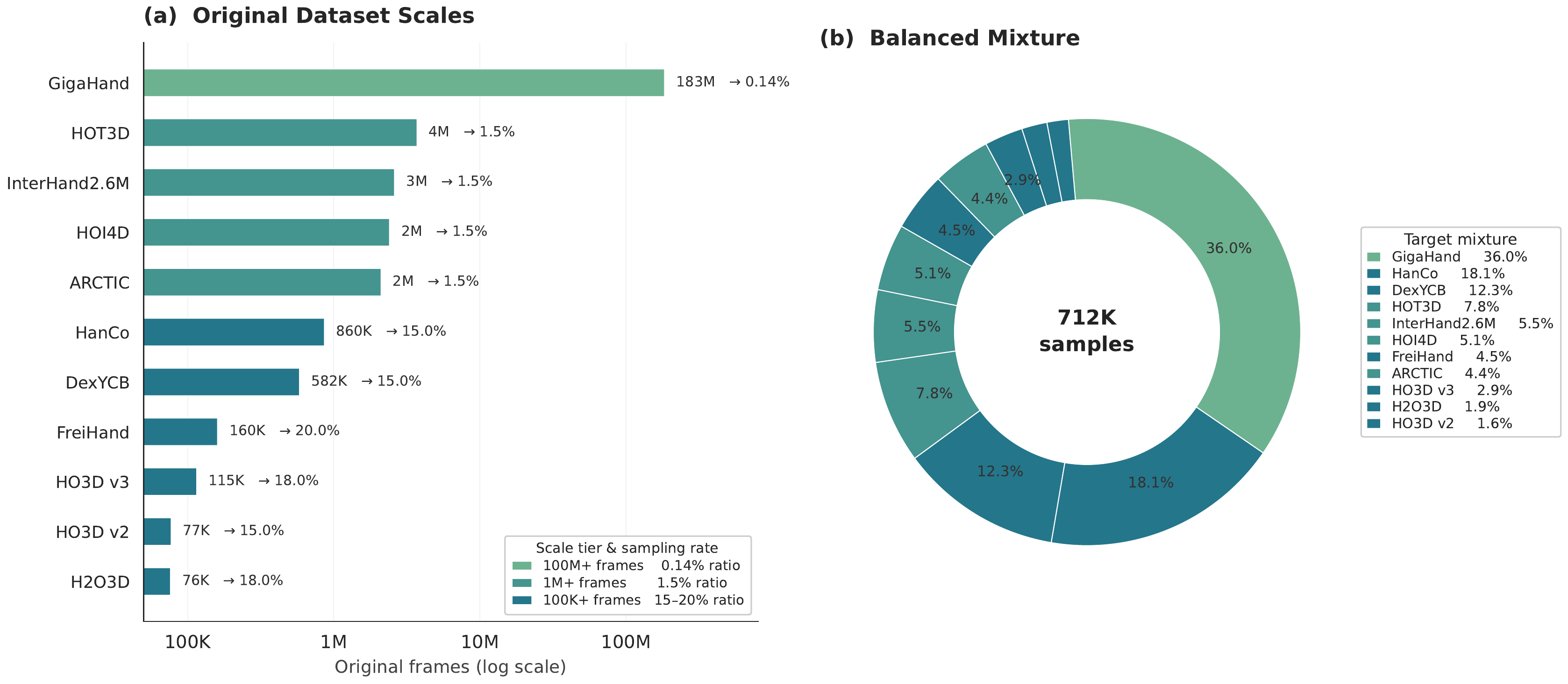}
\put(2.5,37.6){\colorbox{white}{\parbox[c][6pt][c]{0.053\textwidth}{\raggedleft\sffamily\fontsize{4.9}{6}\selectfont GigaHands}}}
\put(88.65,26.8){\colorbox{white}{\parbox[c][5pt][c]{0.048\textwidth}{\sffamily\fontsize{4.4}{5}\selectfont GigaHands}}}
\end{overpic}
\endgroup
\caption{Composition of the \marker{} training data. Left: original frame
counts on a logarithmic scale, with sampling ratios annotated. Right:
dataset proportions after sampling.}
\label{fig:markers2mano_data}
\end{figure*}

We train on hand meshes from 11 public datasets:
GigaHands~\citep{gigahand}, HOT3D~\citep{hot3d}, ARCTIC~\citep{arctic},
HOI4D~\citep{hoi4d}, HanCo~\citep{hanco}, DexYCB~\citep{dexycb},
H2O~\citep{h2o}, InterHand2.6M~\citep{interhand},
FreiHand~\citep{freihand}, HO3Dv3~\citep{ho3d}, and
HO3Dv2~\citep{ho3d}. Together, these datasets contain more than 195 million frames.
To balance the training distribution, we assign each dataset a sampling ratio
based on dataset size: large-scale datasets (100M+ frames, e.g., GigaHands)
at 0.14\%; medium-scale datasets (1M+, e.g., HOT3D, InterHand2.6M) at 1.5\%;
and small-scale datasets (100K+, e.g., FreiHand, HO3D) at 15--20\%. This
yields approximately 0.7M training meshes
(Figure~\ref{fig:markers2mano_data}).

The loss function combines full-mesh supervision with an auxiliary loss on the
16 markers retained by the Transformer:
\begin{equation}
\begin{aligned}
\mathcal{L} ={}&
\frac{1}{V}\sum_{v=1}^{V}
\left\|\mathbf{v}^{\text{pred}}_v-\mathbf{v}^{\text{gt}}_v\right\|_2^2 \\
&+\lambda_{\text{markers}}
\frac{\sum_{n=1}^{16}\alpha_n
\left\|\mathbf{m}^{\text{pred}}_n-\mathbf{m}^{\text{aug}}_n\right\|_2^2}
{\sum_{n=1}^{16}\alpha_n}.
\end{aligned}
\end{equation}
Here, $\mathbf{v}^{\text{pred}}$ and $\mathbf{v}^{\text{gt}}$ denote the
predicted and ground-truth MANO mesh vertices ($V=778$), while
$\mathbf{m}^{\text{pred}}$ and $\mathbf{m}^{\text{aug}}$ denote the predicted
and augmented positions of the 16 retained markers. We use
$\lambda_{\text{markers}}=0.5$ and assign weights $\alpha_n=2.0$ to
fingertip markers and 1.0 to the remaining markers.

\paragraph{Optimization.}
We train \marker{} from scratch for 50 epochs with Adam at a learning rate of $5\times10^{-6}$,
batch size 320, bf16 mixed precision, and gradient clipping at 1.0 on five
GPUs. Validation is performed every epoch, and the checkpoint with the lowest
validation loss is retained.

\paragraph{Data augmentation.}
To improve robustness to motion-capture variation, we apply the following
perturbations to every training batch. Validation uses the same coordinate normalization without augmentation.

\begin{itemize}[leftmargin=1.5em,nosep]
    \item \textbf{Structural}: 
    (1)~Bone-length perturbation---each bone segment is selected independently
    with probability 0.3 and scaled by up to \(\pm 5\%\) of its length,
    simulating soft-tissue artifacts.
    (2)~Random global scaling---after root-centering, all coordinates are multiplied by a factor in \([0.75, 1.25]\).
    \item \textbf{Marker correspondence and placement}: 
    (3)~Marker swap---spatially proximate markers (within 15\,mm) are randomly exchanged with probability 0.3 (capped at three swaps per frame).
    (4)~Neighborhood blending---each marker is blended with a random convex combination of its neighboring mesh vertices, retaining a self-weight of 0.5.
    \item \textbf{Noise}: 
    (5)~Gaussian noise---i.i.d.\ \(\mathcal{N}(0,1\,\text{mm}^2)\) noise is added, with the noise suppressed for a marker with probability 0.1.
    (6)~Marker drift---three markers receive a systematic spatial offset of up to 5\,mm in a random direction.
\end{itemize}

\begin{figure*}[p]
\centering
\begin{overpic}[width=0.80\textwidth]{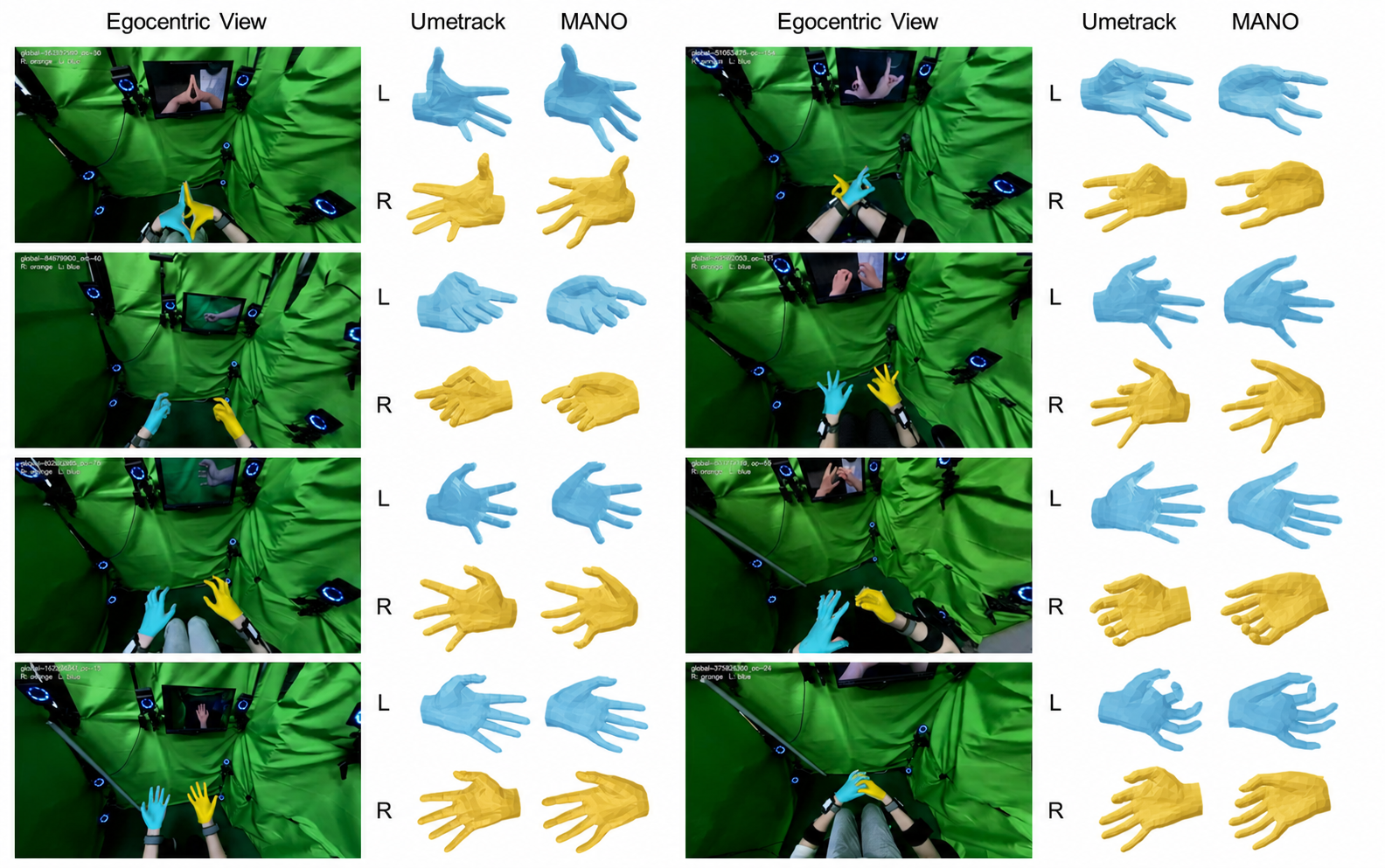}
\put(29.1,60.3){\colorbox{white}{\parbox[c][9pt][c]{0.065\textwidth}{\centering\sffamily\fontsize{7}{8}\selectfont UmeTrack}}}
\put(77.5,60.3){\colorbox{white}{\parbox[c][9pt][c]{0.065\textwidth}{\centering\sffamily\fontsize{7}{8}\selectfont UmeTrack}}}
\end{overpic}
\caption{Qualitative examples of UmeTrack and MANO reconstructions on \dataset{}. The MANO overlays show valid, non-interpolated meshes reconstructed with \marker{} and projected onto egocentric RGB frames using the recorded camera calibration and poses.}
\label{fig:label_quality}
\par\medskip
\includegraphics[width=0.80\textwidth]{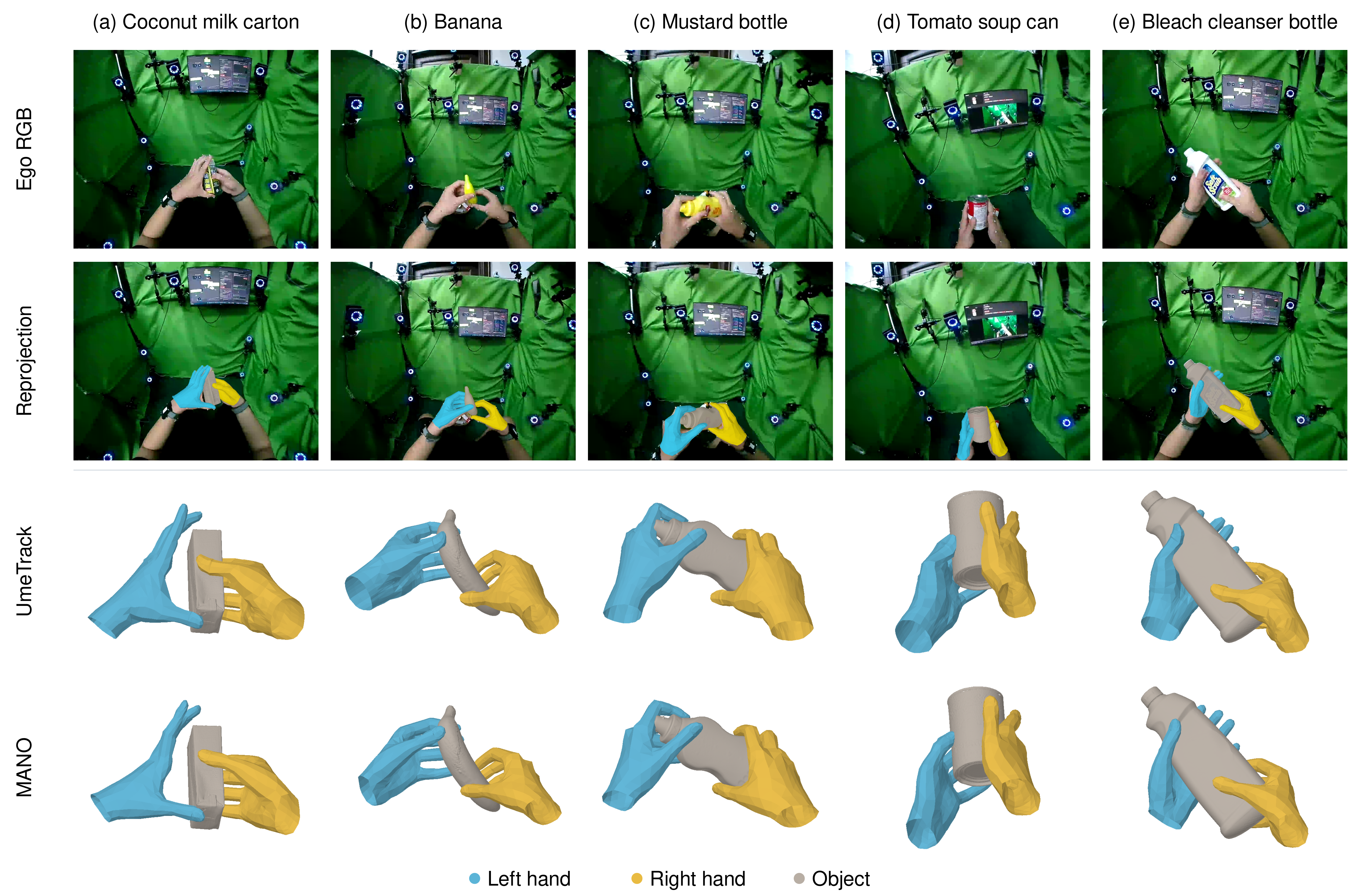}
\caption{Qualitative reconstruction examples from the hand--object interaction
subset of \dataset{}. From top to bottom, the rows show the egocentric RGB frame, a semitransparent
overlay of the reprojected hands and object, a 3D scene with UmeTrack hand
meshes, and a 3D scene with MANO hand meshes. Cyan denotes the left hand, gold the right
hand, and gray the object. Both 3D rows use the same viewpoint, center,
and scale for each example.
Hand labels are valid and non-interpolated; reprojections use the recorded
camera calibration and poses. Objects are positioned using their meshes
and FoundationPose estimates~\citep{foundationposewen2024}.}
\label{fig:hoi_label_quality}
\end{figure*}

We randomly reserve 10\% of the public data for testing. On this test set,
\marker{} achieves a mean per-vertex reconstruction error of 3.2\,mm
relative to the reference meshes. Across \dataset{}, including bare-hand
and HOI recordings, the mean alignment error between captured markers and the corresponding
mesh vertices is 4.3\,mm. We also inspect mesh reprojections on randomly sampled egocentric RGB frames (Figure~\ref{fig:label_quality}).

\renewcommand{\dblfloatpagefraction}{0.8}

\subsection{Estimated MANO Shape Variation}
\label{app:mano_shape_diversity}

We analyze MANO shape estimates for both hands at the original motion-capture
sampling rate, before resampling to match EMG. We exclude invalid estimates
and interpolated frames, leaving 27{,}889{,}846 of 30{,}500{,}174 hand-pose
estimates from 160 hand recordings, where each hand recorded in a session
is counted separately.
For each hand in each session, we compute the median of each of the ten
shape coefficients. We then take the median across sessions for each
participant's left and right hands separately, giving sessions equal weight.

\begin{figure}[t]
\centering
\includegraphics[width=\columnwidth]{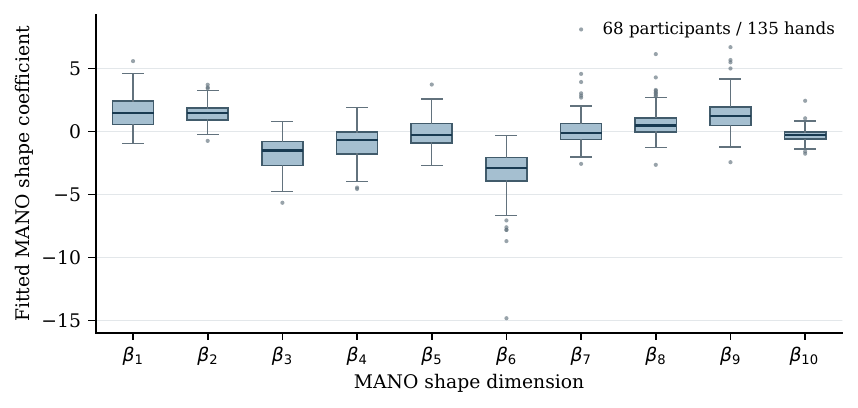}
\caption{MANO shape-coefficient distributions for 135 hands from 68 participants.
Boxes show the interquartile range, lines
show medians, whiskers extend to the most extreme observations within
1.5 interquartile ranges of each box, and points show observations beyond
the whiskers.}
\label{fig:mano_beta_diversity}
\end{figure}

This produces one shape estimate for each of 135 hands from 68 participants:
68 left hands and 67 right hands.
Figure~\ref{fig:mano_beta_diversity} shows the resulting coefficient
distributions. Both hands use the MANO right-hand parameter convention.

\section{Implementation Details}
\label{app:implementation}

\subsection{Training and Evaluation}
\label{app:training_objective}

\paragraph{Training objective.}
On \dataset{}, EMGFormer, vision, and fusion models minimize the masked
angular MAE defined in Eq.~\mainref{eq:benchmark_mae}, computed in radians
for training. EMG-only models are supervised over valid joint angles of
the target hand throughout each window; vision and fusion models,
including temporal variants, are supervised on both hands at the center
frame. A minibatch with no valid targets contributes zero loss.

\paragraph{Inputs and evaluation sampling.}
On \dataset{}, each EMG input contains 4\,s (8{,}000 samples) of
eight-channel data from one wrist. EMGFormer is evaluated on non-overlapping windows. Vision and
fusion use center frames sampled every 400\,ms, with fusion EMG windows
centered on these frames. The additional EMGFormer center-frame evaluation
uses the same grid, retaining centers with valid EMG windows. All windows
and image sequences remain within data partitions, recording segments,
and task boundaries.

\paragraph{EMG baselines on \dataset{}.}
emg2pose, vemg2pose, NeuroPose, SensingDynamics, and \model{} share the
training, validation, and test partitions and predict 22 joint angles
for the target hand. They use the EMG preprocessing described in
\supplementref{app:emg_processing} and the same normalization, with statistics
computed from training data and fixed for validation and testing. All
models are initialized randomly, trained with one seed per model under
the settings in \supplementref{app:training_details}, and selected by
validation MAE. The other four baselines use no data augmentation;
\model{} uses the augmentations described there.

\paragraph{EMG2Pose evaluation.}
\label{app:emg2pose_protocol}
We follow the official EMG2Pose test splits for User, Stage, and
User+Stage generalization. EMGFormer is evaluated on non-overlapping
7,790-sample windows using non-causal context. We compute angular MAE
over valid targets within each user and report the mean and population
standard deviation across users. Results from prior work are reported under
their original evaluation protocols.

\subsection{EMGFormer}
\label{app:training_details}

\paragraph{Optimization.}
All three bidirectional \model{} sizes are trained on \dataset{} for 150 epochs. We use AdamW~\citep{adamw} ($\beta_1=0.9$, $\beta_2=0.999$, weight
decay $10^{-2}$), an initial learning rate of $5\times10^{-4}$ with cosine
annealing to $5\times10^{-6}$, gradient clipping at 1.0, and bf16 mixed
precision.
Training uses four to six NVIDIA RTX 4090 GPUs depending on model size.

\paragraph{Data augmentation.}
All three model sizes use the same training augmentations: random
gain, magnitude warping, baseline drift, power-line noise, time/frequency
masking, additive Gaussian noise, channel masking, mixup, and temporal window
jitter.

\begin{table}[t]
\centering
\caption{Architecture configurations for each \model{} variant. ``Params'' includes the TDS encoder, Transformer decoder, and regression head.}
\label{tab:training_config}
\footnotesize
\setlength{\tabcolsep}{2.0pt}
\renewcommand{\arraystretch}{1.05}
\begin{tabular}{lccccc}
\toprule
Variant & $L$ & $d$ & Heads & $d_{\mathrm{FFN}}$ & Params \\
\midrule
\model{}-S  & 3 & 256 & 4  & 512  & 3.5M \\
\model{}-M & 6 & 256 & 8  & 1024 & 6.6M \\
\model{}-L  & 8 & 384 & 12 & 1536 & 16.3M \\
\bottomrule
\end{tabular}

\end{table}

\paragraph{TDS feature extractor.}
The TDS frontend~\citep{tds} maps each EMG window to a 256-dimensional feature sequence; Table~\ref{tab:training_config} gives the Transformer configurations. The four TDS blocks in
Figures~\mainref{fig:architecture} and~\mainref{fig:fusion_architecture}
comprise two strided 1D convolution blocks followed by two blocks that
combine strided convolution with a TDS encoder. The latter alternate
time-depth separable convolutions with channel-wise fully connected
layers, followed by squeeze-and-excitation~\citep{senet}.
The layer-wise configuration is:
\begin{enumerate}[leftmargin=1.5em,nosep]
    \item TDS Block 1: 1D convolution, $8 \to 256$, kernel 11, stride 5
    \item TDS Block 2: 1D convolution, $256 \to 256$, kernel 5, stride 2
    \item TDS Block 3: input convolution kernel 9, stride 5; 1 block of TDS convolution (kernel width 5, channels 8, feature width 32)
    \item TDS Block 4: input convolution kernel 3, stride 1; 1 block of TDS convolution (kernel width 3, channels 8, feature width 32)
\end{enumerate}
The bidirectional models use non-causal self-attention over the full feature
sequence. The causal variants are described in
\supplementref{app:emg_model_causal}.

\subsection{Vision and Fusion Models}
\label{app:training_details_vision}

Vision and fusion models use AdamW~\citep{adamw}, cosine annealing, and
bf16 mixed precision.

\paragraph{Vision baseline.}
For an image feature $\mathbf{z}_t\in\mathbb{R}^{d_v}$, the bimanual
regression head consists of $\mathrm{Linear}(d_v,512)$, ReLU,
dropout with probability 0.1, and $\mathrm{Linear}(512,44)$. Its output is
reshaped into left and right 22-angle vectors.
Table~\ref{tab:vision_config} lists the backbone dimensions and training settings.

\begin{table}[t]
\centering
\caption{Vision backbone dimensions and training settings. All backbones
use the same regression head architecture.}
\label{tab:vision_config}
\footnotesize
\setlength{\tabcolsep}{3.0pt}
\renewcommand{\arraystretch}{1.05}
\begin{tabular}{lccccc}
\toprule
Backbone & Params & $d_{\text{backbone}}$ & LR & Batch size & Epochs \\
\midrule
ResNet-18  & 11.5M  & 512  & $10^{-3}$ & 900 & 150 \\
ResNet-50  & 23.5M  & 2048 & $10^{-3}$ & 400 & 150 \\
ResNet-152 & 58.2M  & 2048 & $10^{-3}$ & 180 & 150 \\
ViT-S/14   & 21.9M  & 384  & $10^{-4}$ & 360 & 150 \\
ViT-B/14   & 86.2M  & 768  & $10^{-4}$ & 160 & 150 \\
ViT-L/14   & 303.8M & 1024 & $10^{-4}$ & 50  & 150 \\
\bottomrule
\end{tabular}

\end{table}

\paragraph{Fusion baseline.}
Fusion uses the shared \model{}-S encoder and decoder, followed
by temporal attention pooling. The pooling module scores each
256-dimensional decoder token with a shared
$256\to64\to1$ MLP with Tanh, normalizes scores with a temporal softmax,
and computes their weighted sum.

The visual projection $P_v$ is a shared linear map from $d_v$ to 256.
Concatenating the pooled EMG feature, projected image feature, and one-hot
hand identity gives 514 dimensions. The shared projection $G$ comprises
$\mathrm{Linear}(514,512)$, GELU, dropout (0.1), and
$\mathrm{Linear}(512,256)$. Each independent head $H_h$ comprises
$\mathrm{Linear}(256,512)$, ReLU, dropout (0.1), and
$\mathrm{Linear}(512,22)$. The final-layer weights are sampled from a zero-mean Gaussian with a
standard deviation of $10^{-5}$, and the biases are initialized to zero.
This yields the near-zero residual initialization in
Eq.~\mainref{eq:bimanual_residual_fusion}. When a hand's EMG window is unavailable,
its correction is set to zero. Both pretrained branches and all added
layers are fine-tuned jointly.

\paragraph{Direct fusion ablation.}
Direct and residual fusion use the same pretrained vision and EMG models,
data, optimizer, learning-rate schedule, training budget, and checkpoint
selection rule. They differ only in how they predict joint angles from the fused features:
direct fusion predicts the angles directly, whereas residual fusion
predicts a correction to the vision estimate.

\paragraph{Temporal vision baseline.}
\label{app:temporal_visual_control}
For the ViT-S temporal models, one shared encoder extracts
$\mathbf{z}_k\in\mathbb{R}^{384}$ at nominal times
$t+0.25k\,\mathrm{s}$, $k=-8,\ldots,8$. A linear map
$A:\mathbb{R}^{384}\to\mathbb{R}^{1024}$ and learned temporal positional embeddings
$\mathbf{p}_k$ form 17 tokens. A single pre-normalized Transformer layer
$\mathcal{T}$ uses eight attention heads, a 5504-dimensional feed-forward
hidden layer with GELU, and dropout 0.1. The center token refines the
original center feature through $B:\mathbb{R}^{1024}\to\mathbb{R}^{384}$:
\[
\begin{aligned}
\mathbf{U}_t &= [A\mathbf{z}_k+\mathbf{p}_k]_{k=-8}^{8},\\
\widetilde{\mathbf{z}}_t
 &= \mathbf{z}_0+B\!\left(\mathcal{T}(\mathbf{U}_t)_0\right).
\end{aligned}
\]
The output projection $B$ is initialized to zero. The same 44-output regression head maps $\widetilde{\mathbf{z}}_t$ to
both hands' joint angles at the center frame.
Missing context frames are replaced with the center image and masked as
attention keys; the center frame is always valid.
Temporal + EMG substitutes $\widetilde{\mathbf{z}}_t$ for
$\mathbf{z}_t$ in both the visual head and $P_v$ of
Eq.~\mainref{eq:bimanual_residual_fusion}; the remaining fusion components
are unchanged.

\FloatBarrier
\end{document}